\pdfoutput=1
\documentclass[aps,prx,reprint,superscriptaddress,nofootinbib,
               amsmath,amssymb,floatfix,longbibliography]{revtex4-2}

\usepackage{graphicx}
\usepackage{booktabs}
\usepackage{microtype}
\usepackage[dvipsnames]{xcolor}
\usepackage{url}
\usepackage[colorlinks=true,linkcolor=RoyalBlue,citecolor=RoyalBlue,
            urlcolor=RoyalBlue]{hyperref}

\newcommand{\btheta}{\boldsymbol{\theta}}
\newcommand{\Sset}{\mathcal{S}}

\begin{document}

\title{Dynamical Batching: How Class-Cycling Training Shapes the Internal Structure of Neural Networks}

\author{Nicol\'as Ratier-Werbin}
\altaffiliation[]{These authors contributed equally to this work. Contact authors: \href{mailto:nratier@ucm.es}{nratier@ucm.es}, \href{mailto:samuel.lozano@ucm.es}{samuel.lozano@ucm.es}}
\affiliation{Departamento de Estructura de la Materia, F\'isica T\'ermica y Electr\'onica, Universidad Complutense de Madrid, 28040 Madrid, Spain}

\author{Samuel Lozano}
\altaffiliation[]{These authors contributed equally to this work. Contact authors: \href{mailto:nratier@ucm.es}{nratier@ucm.es}, \href{mailto:samuel.lozano@ucm.es}{samuel.lozano@ucm.es}}
\affiliation{Departamento de Estructura de la Materia, F\'isica T\'ermica y Electr\'onica, Universidad Complutense de Madrid, 28040 Madrid, Spain}
\affiliation{Grupo Interdisciplinar de Sistemas Complejos, Universidad Complutense de Madrid, 28040 Madrid, Spain}

\author{Miguel Ruiz-Garc\'ia}
\affiliation{Departamento de Estructura de la Materia, F\'isica T\'ermica y Electr\'onica, Universidad Complutense de Madrid, 28040 Madrid, Spain}
\affiliation{Grupo Interdisciplinar de Sistemas Complejos, Universidad Complutense de Madrid, 28040 Madrid, Spain}

\date{\today}

\begin{abstract}
We investigate whether cyclically varying the class composition of training batches can modify both the performance and the internal organization of neural networks. Motivated by nonequilibrium systems, in which cycling between related sub-goals can make switching between their solutions increasingly efficient, we introduce \emph{Dynamical Batching} (DB), a sampling protocol that periodically emphasizes each class while leaving the loss and the optimizer unchanged, and compare it with Uniform Batching (UB), in which every mini-batch has equal class proportions. In the regime where standard training fails, DB changes the outcome: on a three-arm spiral, a capacity-limited MLP that plateaus far below perfect accuracy under UB fits the training data under DB. Tracking individual weights, we find that under DB the change between class-specialized states is carried by a smaller set of weights than under UB, and that this set is markedly more persistent from cycle to cycle, on every synthetic dataset we test and in a wider network that fits the data under both protocols. This organization is a signature of the protocol rather than a predictor of its benefit: it appears equally where DB leaves accuracy unchanged. On CIFAR-10, the same protocol modestly improves the test accuracy of a channel-narrowed ResNet-10, a standard residual architecture, and of a second capacity-limited convolutional network. These results show that how class emphasis is distributed in time can shape both the solution a network reaches and the internal structure of the network.
\end{abstract}

\maketitle

\section{Introduction}

Neural networks are typically optimized under a fixed training objective, yet the trajectory followed to optimize that objective is itself part of the learning process. Choices such as the ordering of examples, curriculum, learning-rate schedule, optimizer, and batch construction can change which solutions are reached even when the final objective is unchanged \cite{bengio2009curriculum,smith2017cyclical,neyshabur2017implicit,cohen2021,cohen2024understanding}. This raises a broader question: how does a training protocol change not only which function a network learns, but also where in parameter space different parts of that function are implemented?

The question is particularly interesting for tasks that contain several related sub-problems. In a multiclass classifier, for example, a single network must implement class-specific decision rules while sharing a finite parameter budget. More generally, multitask learning asks a shared model to solve several objectives whose gradients may not point in the same direction. When task gradients conflict, optimizing their sum can produce inefficient or destructive interference, motivating methods that explicitly rebalance, project, or otherwise manipulate task gradients \cite{sener2018multi,yu2020gradient,chen2018gradnorm}. Continual-learning methods approach a related problem from the opposite temporal regime: when tasks are presented sequentially, parameter isolation, masks, pruning, or architectural expansion can protect task-specific computations from interference \cite{kirkpatrick2017forgetting,serra2018hat,mallya2018piggyback,mallya2018packnet,rusu2016progressive,ruiz2022model}, and whether tasks are presented in blocks or interleaved changes what is learned \cite{flesch2018comparing}. These approaches demonstrate that \emph{which parameters are allowed to change} is consequential, but they typically impose parameter isolation explicitly or address tasks one after another.

Here we study a different regime: the architecture and the objective are unchanged, but optimization pressure is deliberately cycled between related sub-problems. The protocol is closely related to  \emph{dynamical loss functions} \cite{ruizgarcia2021,lavin2025dynamical},
an approach motivated by oscillating strategies in biological systems, materials science and physics \cite{murugan2019bioinspired,murugan2021roadmap,mustonen2009fitness,cairns2022strong,falk2023, ruiz2019tuning}, 
and distinct from learned, teacher-driven loss schedules \cite{wu2018dynamicloss}. A dynamical loss periodically changes the class weights of the cross-entropy loss, and this can rescue under-parametrized networks that fail under a static loss and can improve generalization in over-parametrized ones \cite{ruizgarcia2021,lavin2025dynamical}. This process can be analyzed through Hessian and neural-tangent-kernel dynamics \cite{ruizgarcia2021}. Our questions are complementary: can a similar temporal forcing, implemented through mini-batch composition, rescue under-parametrized networks that fail under Uniform Batching (UB), in which every mini-batch has equal class proportions? And does it reorganize the parameters so that switching between class-specific solutions becomes more economical? Acting on batch composition rather than on the loss leaves the per-example loss and the optimizer untouched, so the protocol can be dropped into any training pipeline without modification.

This question connects to a broader literature on modularity. Modularity is commonly understood as an organization in which subsets of a system support distinguishable functions while interacting relatively weakly with other subsets. In neural networks, modularity can be defined structurally through connectivity \cite{filan2021clusterability} or functionally through subsets of weights required for particular computations \cite{csordas2021neural,hod2021modularity}. Standard networks can exhibit some specialization without being explicitly modular, but the extent to which useful functional modules emerge from ordinary training remains an active question. In particular, \citet{csordas2021neural} found that common architectures often fail to reuse the same submodules across related functions, while work on superposition has emphasized that finite representational capacity can instead lead multiple features to share neurons or directions \cite{elhage2022superposition,scherlis2022polysemanticity}. We therefore do not assume that a network becomes modular in the graph-theoretic sense. Instead, we ask whether the parameter changes associated with switching between sub-problems become localized: a weaker notion than modularity, but a directly measurable one.

Our central hypothesis is that periodically emphasizing each class concentrates the class-to-class change of the network onto a small, stable set of weights. If so, moving from one class-specialized state to another should involve fewer weights, largely the same ones every time, and a shorter distance in parameter space. We test this prediction with three complementary measurements: the size of the \emph{switching set}, the smallest fraction of weights that carries most of the class-to-class change; its \emph{persistence}, the overlap between the switching sets of consecutive cycles; and the \emph{switching distance} between class-specialized states.

We first map the operating regime of Dynamical Batching (DB) on a three-arm spiral, where a capacity-constrained MLP that plateaus under UB reaches near-perfect training accuracy under a suitable amplitude--period combination. We then follow the same training runs over time: the class-to-class change concentrates on a small set of weights that persists from cycle to cycle, while the distance between class-specialized states decreases, although the drift of ordinary training also decreases. Finally, we test whether the accuracy effect extends beyond the toy setting with a channel-narrowed ResNet-10 on CIFAR-10 and three further convolutional architectures. Our contributions are:
\begin{enumerate}
    \item \textbf{A batch-level protocol that rescues a capacity-limited network.} Without changing the loss, optimizer or architecture, DB lets a one-hidden-layer MLP that plateaus under UB fit the spiral; a grid over amplitude and period shows a broad region of gain bordered by a failure region.
    \item \textbf{Weight-level measures of class switching.} We define a drift-corrected switching set, its persistence, and a switching distance, and track them against a UB control that uses identical bookkeeping.
    \item \textbf{Persistent, localized switching as a signature of the protocol.} Under DB, the switching set is smaller and approximately twice as persistent as under UB across all five synthetic datasets, and the same pattern holds for a wider network that fits the data under both protocols.
    \item \textbf{Evidence beyond the toy setting.} On CIFAR-10, DB modestly improves the test accuracy of a channel-narrowed ResNet-10, acting as a regularizer; a plain CNN with a narrow fully connected bottleneck shows a similar gain, while two wider networks show none beyond seed-to-seed noise.
\end{enumerate}

\section{Dynamical Batching}
\label{sec:setup}

\subsection{Protocol}

\begin{figure*}[tb]
    \centering
    \includegraphics[width=\textwidth]{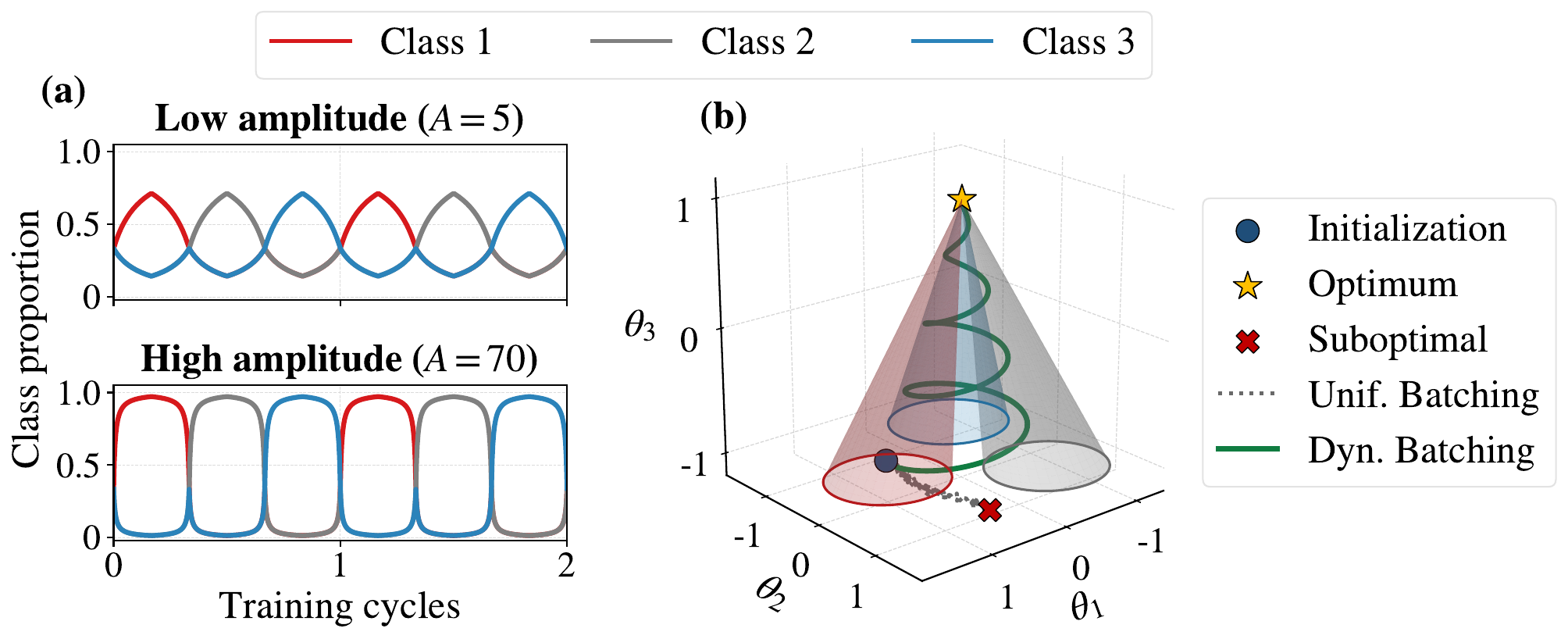}
    \caption{\textbf{Learning by Dynamical Batching (DB).} \textbf{(a)} Class proportion $p_i(t)$, the fraction of each mini-batch drawn from class $i$, over two training cycles ($C=3$). At low amplitude (top, $A=5$) every batch still contains all three classes: the focal-class proportion rises from $1/3$ to $0.71$, while the other two fall to $0.14$. At high amplitude (bottom, $A=70$) the emphasized class dominates the batch, its proportion saturating near $1$, while the other two nearly vanish. A cycle is $C$ consecutive periods, one sweep through the classes. \textbf{(b)} Schematic of the parameter space (illustrative, not data). Each class-specialized objective is satisfied in an extended, cone-shaped region (one cone per class, colored as in (a)), and the cones meet only near a common apex (Optimum, star). The shape only encodes that each class alone is satisfied in a large region of parameter space, while all classes together are satisfied only in a much smaller region. In the under-parametrized regime, Uniform Batching (UB, dotted gray) moves from initialization (circle) to a point that satisfies no class fully (Suboptimal, cross). DB (solid green) alternates between the cones, each period pulling the trajectory toward one class's region before the next pulls it toward another.}
\label{fig:protocol}
\end{figure*}

Figure~\ref{fig:protocol} shows the DB protocol schematically. Each mini-batch is built with a time-dependent number of samples from each class. Panel (a) shows the fraction of samples drawn from each class over time, at low and high amplitude. Panel (b) is a cartoon of parameter space: the three colored cones represent regions where the class of that color is correctly classified, and the lines display training trajectories.

Our working hypothesis is that under UB a capacity-limited network reaches a compromise point that satisfies no class fully, whereas DB forces the network to visit a different region of parameter space each time a different class is emphasized and, as training progresses, the trajectory approaches a region where all classes are classified correctly. Moving from one class-emphasized region to another should then require increasingly small changes in parameter space, which we test directly in Sec.~\ref{sec:cost}.

To integrate the oscillations into standard optimizers without altering the loss function (standard cross-entropy optimized with SGD or Adam), we implement the protocol as a dataset sampling schedule. For a dataset of $C$ classes and period $T$, one focal class $c^*(t) = \lfloor t/T \rfloor \bmod C$ is emphasized during its window. Its relative sampling weight $g(t_T)$ follows a triangular ramp of amplitude $A\ge 1$:
\begin{equation}
    g(t_T)=\begin{cases}1+m\,t_T, & 0\le t_T\le T/2,\\ 2A-1-m\,t_T, & T/2<t_T< T,\end{cases}
\label{eq:ramp}
\end{equation}
where $t_T = t\bmod T$ and $m=2(A-1)/T$. We call $C$ consecutive periods a \emph{cycle}. The class proportions are $p_{c^*}(t) = \frac{g(t_T)}{g(t_T) + C - 1}$ for the focal class and $p_j(t) = \frac{1}{g(t_T) + C - 1}$ for each of the other $C-1$ classes (Fig.~\ref{fig:protocol}a). $A=1$ recovers UB.

Each mini-batch of nominal size $B$ is built by deterministic stratification: class $i$ contributes
\begin{equation}
    n_i(t) = \lfloor p_i(t) B \rfloor
    \label{eq:collapse}
\end{equation}
examples, drawn with replacement within the class. The class composition is therefore fixed at every step; only the choice of samples within each class is random.\footnote{Because of the flooring, the realized batch size $\sum_i n_i(t)$ can fall short of $B$ by fewer than $C$ samples. This happens even under UB: $B=50$ and $C=3$ give $3\times16=48$. At large amplitudes, flooring also removes the non-focal classes entirely, since $n_{j\neq c^*}=0$ whenever $g(t_T) > B-C+1$.} In the final $5\%$ of training steps all batches are uniform, so that every run ends under the same objective.

\subsection{Experimental setup}
\label{sec:exp_setup}

To understand the learning dynamics in detail, we use a three-arm spiral, a multi-class variant of the classic two-spirals benchmark \cite{lang1988learning}, with $100$ training points per class and another $100$ test points per class, generated by the same procedure with a different random seed; each training seed draws a new dataset and a new initialization (Appendix~\ref{sm:toy}). We train two one-hidden-layer MLPs with ReLU activations and Glorot initialization \cite{glorot2010understanding}: a small network with $H=50$ hidden units ($303$ parameters; under-parametrized for this task) and a large one with $H=500$ ($3{,}003$ parameters; over-parametrized). All main-text MLP experiments use $125{,}000$ training steps, batch size $B=50$ and Adam \cite{kingma2015adam} (learning rate $0.002$, $\beta=(0.9, 0.999)$, $\epsilon=10^{-8}$), with zero weight decay and no gradient clipping.

\section{Dynamical Batching succeeds where Uniform Batching plateaus}
\label{sec:regime}

\begin{figure*}[tb]
    \centering
    \includegraphics[width=\textwidth]{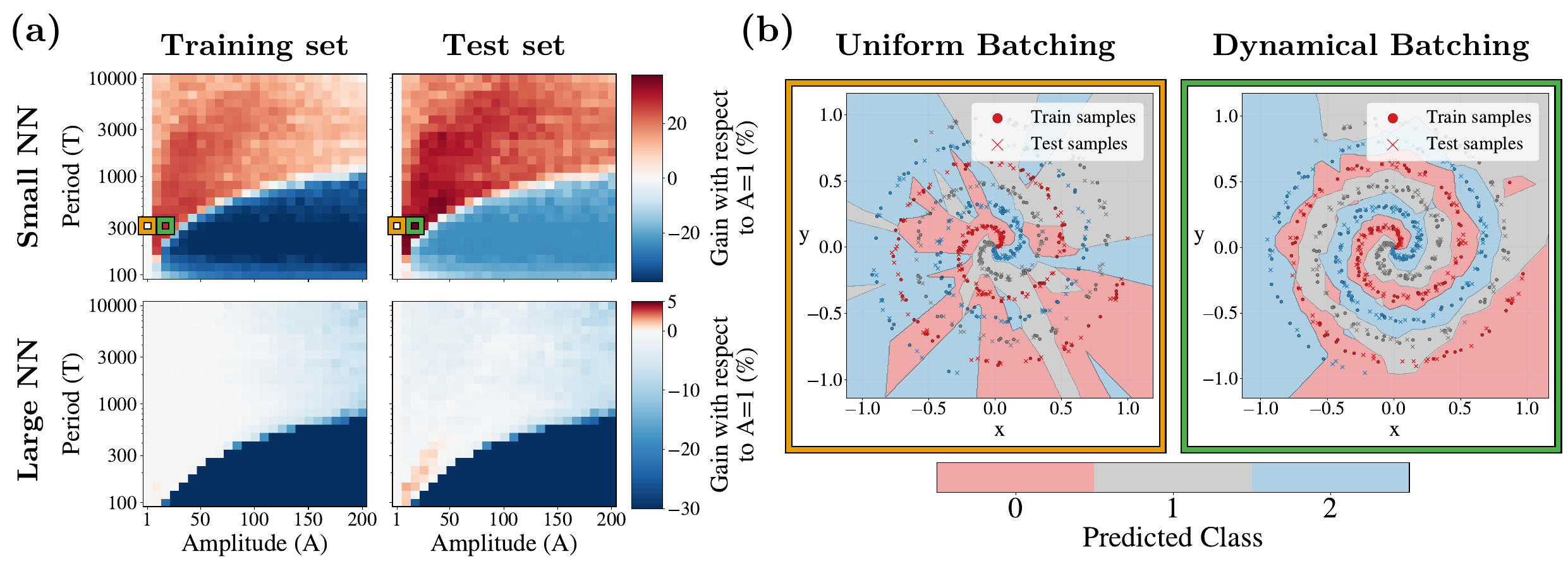}
    \caption{\textbf{Dynamical Batching (DB) on the three-arm spiral.} \textbf{(a)} Accuracy gain in percentage points over the Uniform Batching (UB, $A=1$) run with the same period $T$, across oscillation amplitude $A$ and period $T$, for the small ($H=50$) and large ($H=500$) networks; training and test gain shown separately. Color scales differ between rows. Each cell averages 5 seeds; the yellow and green boxes mark the UB baseline ($A=1$) and the best DB configuration, $(A,T)=(20,320)$, used in (b) and in Sec.~\ref{sec:cost}. For the small network, a broad region of positive gain reaches $+37.87$ points in test accuracy, next to a failure region at large $A$ and small $T$. The large network already reaches $99.00\%$ train and $95.93\%$ test accuracy under UB in this grid, and DB improves test accuracy slightly (up to $+1.74$ points) in a narrow band at low amplitude, just above the failure region. \textbf{(b)} Decision boundaries for a single UB run ($A=1$) and a single DB run at $(A,T)=(20,320)$: the network is evaluated on a grid and shaded by predicted class, with training (circles) and test (crosses) samples colored by true label. UB produces a fragmented boundary that does not follow the arms (test accuracy $56.60\%$); DB recovers the spiral structure (test accuracy $94.47\%$).}
\label{fig:phase}
\end{figure*}

We ask whether DB can improve learning, and under which conditions. Figure~\ref{fig:phase} maps the accuracy gain over the $A=1$ (UB) baseline across amplitude and period (5 seeds per cell). A broad region of positive test-accuracy gain emerges for moderate amplitudes and periods, where the oscillating emphasis lets the network leave the poor solution reached under UB. At large amplitudes combined with short periods a failure region appears, consistent with the emphasis shifting faster than the network can consolidate. For the small network ($H=50$), the best cell of the grid, $(A,T)=(20,320)$, improves test accuracy by $+37.9$ points over the grid's UB baseline. Re-running this configuration and its UB control with 20 seeds each gives $70.5\pm8.9\%$ training and $59.2\pm7.7\%$ test accuracy under UB, against $99.9\pm0.3\%$ and $95.1\pm1.3\%$ under DB (Table~\ref{tab:switching_final}). The UB shortfall is not a matter of slower convergence within our training budget: UB training accuracy gains less than $10$ points over the last two thirds of training and remains far below DB throughout (Appendix~\ref{sm:curves}, Fig.~\ref{fig:sm_accuracy}a), and doubling the budget to $250{,}000$ steps raises it only to $75.3\pm11.3\%$ (Table~\ref{tab:sm-long}). For the large network ($H=500$), UB already fits the training set, and DB improves test accuracy only slightly, by at most $+1.7$ points in the grid, at $(A,T)=(10,150)$; with 20 seeds, test accuracy is $96.6\pm1.5\%$ under UB and $98.2\pm1.2\%$ under DB.

Why does UB plateau when the same network can fit the data under DB? One candidate is a gradient-conflict plateau, related to the conflicting-gradient problem of multi-task learning \cite{yu2020gradient}. When per-class gradients disagree, the balanced gradient $\frac{1}{C}\sum_i\nabla\mathcal{L}_i$ can vanish at a point where no individual class is satisfied ($\nabla\mathcal{L}_i\neq0$ for every $i$), so the mean step of UB is zero there, whereas a class-focused batch follows a single, undiluted $\nabla\mathcal{L}_i$. If $\sum_i\nabla\mathcal{L}_i=0$ and the per-class gradients have equal norm, their mean pairwise cosine must equal $-1/(C-1)$, i.e.\ an angle of $120^\circ$ for $C=3$ (Appendix~\ref{sm:gradients}). We measured this cosine for gradients computed on fixed, single-class probe sets, with all $C$ gradients evaluated at the same parameters (Appendix~\ref{sm:gradients}, Fig.~\ref{fig:sm_gradients}a,b). Under UB it sits at or close to $-0.5$ throughout training, both in the small network that plateaus near $71\%$ training accuracy and in the large network that fits the data. Under DB, evaluated at the class-emphasized states, the class gradients remain strongly opposed but do not cancel (cosine from $\approx-0.39$ early in training to $\approx-0.45$ after $\approx40$--$50$ cycles). Two conclusions follow. First, because the cancelling arrangement is equally present when UB succeeds, it cannot by itself explain the plateau of the small network: it indicates only that UB training stays close to a stationary point of the balanced loss, which is poor in the small network and good in the large one. Second, DB does not remove the conflict between classes; it changes which combination of the conflicting gradients the optimizer follows. The DB batch weights the classes unequally, so its expected gradient is $\sum_i p_i(t)\nabla\mathcal{L}_i$ with $p_{c^*}(t)>1/C$. Even at a point where the balanced sum vanishes, this step equals $\sum_i\big(p_i(t)-\tfrac{1}{C}\big)\nabla\mathcal{L}_i$, which is generically nonzero: the network keeps moving where UB would stall. We regard this as a hypothesis for how DB leaves the poor solution, not as its explanation.

\section{Class switching localizes onto a small, persistent set of weights}
\label{sec:cost}

\label{sec:measures}

We quantify how the protocol organizes the weights with three measures. Let $\btheta_i(c)$ be the average of the parameter vector over the $T/2$ training steps in the central half, $[T/4, 3T/4)$, of class $i$'s period in cycle $c$, where class $i$ dominates the batch.

To ask \emph{which} weights change when the emphasis moves from one class to the next, we measure for each weight $k$ how much it varies across the $C$ class-focused periods of a cycle: $a_k(c)$, the standard deviation of $\theta_{i,k}(c)$ across classes $i$, after removing the part of that variation that is simply steady drift across the consecutive class windows (Appendix~\ref{sm:switch}, Eq.~\eqref{eq:oscillation}). The \emph{switching set} $\Sset(c)$ is the smallest set of weights that together carry $80\%$ of the total variation $\sum_k a_k^2(c)$; all other weights form the \emph{scaffold}. We report its size as a fraction of all $n$ weights, $|\Sset(c)|/n$. The ordering between protocols is unchanged for thresholds of $50\%$ and $90\%$ (Appendix~\ref{sm:switch}, Fig.~\ref{fig:sm_spiral}d).

Whether the switch is carried by the same weights every cycle is measured by the Jaccard overlap of consecutive switching sets, 
\begin{equation}
    J(c)=\frac{|\Sset(c)\cap\Sset(c-1)|}{|\Sset(c)\cup\Sset(c-1)|},
\end{equation}
the number of weights that are switching weights in both cycles, divided by the number that are switching weights in at least one of them.

The simplest measure of how far the network moves when it switches classes is the \emph{switching distance}, the mean distance over the $C(C-1)/2$ class pairs,
\begin{equation}
    d(c)= \frac{2}{C(C-1)}\sum_{i>j} \lVert\btheta_i(c)-\btheta_j(c)\rVert_2 .
    \label{eq:distance}
\end{equation}

UB uses the same schedule and the same bookkeeping, but with $A=1$, so class sampling stays uniform. $\btheta_i(c)$ is then not a class-emphasized state but the average parameter vector over the corresponding window of ordinary training, and its oscillation, switching set and distance measure the fluctuations and drift of ordinary training over the same windows. UB is therefore the natural control for \emph{how concentrated} and \emph{how stable} the motion is, but not for \emph{how large} it is: DB moves between genuinely different class solutions and UB does not, so the distance between class states is expected to be larger under DB.

\begin{figure*}[tb]
    \centering
    \includegraphics[width=\textwidth]{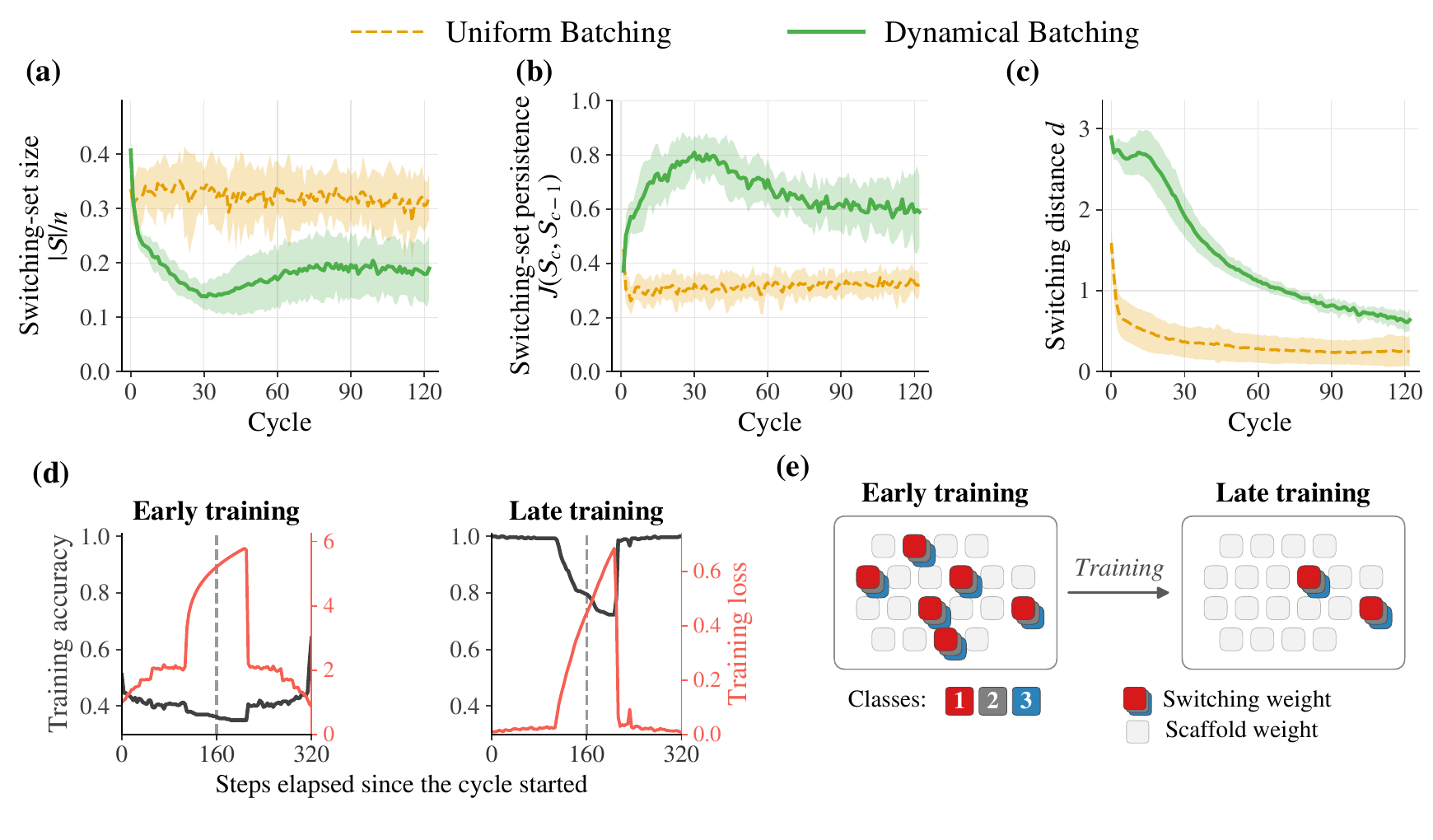}
    \caption{\textbf{Class switching concentrates on a small, persistent set of weights, and the distance between class states shrinks over training.} Spiral, small network, $(A,T)=(20,320)$, $B=50$. One cycle is $3\times320=960$ steps, giving $130$ cycles, of which the first $122$ are dynamical and the rest fall in the uniform final $5\%$ and are excluded from the analysis. Solid green: Dynamical Batching (DB); dashed amber: Uniform Batching (UB, $A=1$). \textbf{(a)} Switching-set size $|\Sset|/n$: the fraction of all weights needed to carry $80\%$ of the class-to-class variation. \textbf{(b)} Persistence of the switching set: the fraction of weights switching in either cycle that switch in both. \textbf{(c)} Switching distance $d$ (Eq.~\eqref{eq:distance}), the mean of $\lVert\btheta_i-\btheta_j\rVert_2$ over the three class pairs. \textbf{(d)} Full-dataset training accuracy (black, left axis) and training loss (coral red, right axis; note the different loss scales) over one focal-class period ($T=320$ steps; the horizontal axis counts steps from the start of that period), early and late in one DB run. The dashed vertical line marks the middle of the period ($t=T/2=160$), where the focal class is most emphasized. The step-like changes of the loss coincide with the steps at which the number of samples of each non-focal class per batch changes (Eq.~\eqref{eq:collapse}). \textbf{(e)} Weight-level schematic (no data): early in training the class-to-class change is spread over many weights and redrawn every cycle; late in training it is carried by a small set of weights (highlighted) that is largely the same in every cycle, while the remaining scaffold barely oscillates. In (a)--(c), each point is the mean over 20 seeds and bands are $\pm1$ s.d.\ across seeds.}
\label{fig:cost}
\end{figure*}

\subsection{Results}

Throughout this section the small network is trained at $(A,T)=(20,320)$,\footnote{For $(A,B,C)=(20,50,3)$ the realized batch size is $48$--$50$, and the focal class makes up $84$--$92\%$ of the batch in the central half of its window.} and the UB control is the identical run with $A=1$.

By the end of training, switching between classes under DB is carried by fewer weights than the fluctuations of UB training, and largely by the same weights every cycle (Fig.~\ref{fig:cost}a,b, Table~\ref{tab:switching_final}). Under DB, $80\%$ of the class-to-class variation is carried by only $14\%$ of the weights after $\approx30$ cycles and by $19\pm6\%$ at the end of training; under UB, the same fraction of the ordinary training fluctuations requires about a third of the weights throughout ($32\pm4\%$ at the end; Fig.~\ref{fig:cost}a). The participation ratio of $\btheta_i-\btheta_j$, a threshold-free measure of the same concentration, agrees in direction but separates the protocols only late in training (Appendix~\ref{sm:switch}, Fig.~\ref{fig:sm_spiral}c). A small switching set could still be redrawn in every cycle, but it is not: consecutive switching sets overlap with a Jaccard index that peaks near $0.8$ around cycle $30$ and ends at $0.59\pm0.13$ under DB, against $0.33\pm0.06$ under UB (Fig.~\ref{fig:cost}b). Both measures are thus most favorable around cycle $30$ and relax partially afterwards, while staying well separated from UB. At the same time, the distance between class-emphasized states falls about $4.5$-fold, from $2.9$ to $0.64\pm0.14$ (Fig.~\ref{fig:cost}c). This decrease is not by itself specific to DB: the drift of ordinary training over the same windows also falls, from $\approx1.6$ to $0.24\pm0.19$, mostly within the first few cycles, and the DB distance remains above it, as it must. Relative to that drift, the DB distance still shrinks, from roughly five times the UB value around cycle $10$ to $2.7$ times at the end. What is specific to DB is that the class states remain distinct while the change between them is carried by a small, persistent set of weights. The three quantities are complementary: a network could switch through a small set of weights that is redrawn every cycle, or through a stable set that moves ever farther. Figure~\ref{fig:cost}e sketches the picture they give together.

\begin{table*}[tb]
\centering
\caption{Spiral, small network, $(A,T)=(20,320)$, mean $\pm$ s.d.\ over 20 seeds. Accuracies are measured at the end of training (after the uniform final phase); switching measures at the last cycle of each protocol's regime (the last dynamical cycle for DB, the last cycle for UB). $|\Sset|/n$: switching-set size; $J$: persistence of the switching set; $d$: switching distance (Eq.~\eqref{eq:distance}).}
\label{tab:switching_final}
\begin{tabular}{lccccc}
\toprule
Method & Train acc.\ (\%) & Test acc.\ (\%) & $|\Sset|/n$ & Persistence $J$ & $d$ \\
\midrule 
UB & $70.5\pm8.9$ & $59.2\pm7.7$ & $0.32\pm0.04$ & $0.33\pm0.06$ & $0.24\pm0.19$ \\
DB & $99.9\pm0.3$ & $95.1\pm1.3$ & $0.19\pm0.06$ & $0.59\pm0.13$ & $0.64\pm0.14$ \\
\bottomrule
\end{tabular}%
\end{table*}

This organization is not a by-product of the accuracy gap of the small network. In the large network ($H=500$, $(A,T)=(10,150)$), both protocols fit the training set (mean training accuracy $\geq99.9\%$), and the first two observations hold more strongly: under DB the switching set shrinks to $3\pm3\%$ of the weights, against $15\pm6\%$ under UB, and its persistence reaches $0.70\pm0.20$, against $0.34\pm0.14$. The distance between class states again falls, from $3.6$ to $0.18\pm0.12$, but here it closely tracks the drift of ordinary training throughout and ends close to it ($0.13\pm0.11$; Appendix~\ref{sm:large}, Fig.~\ref{fig:sm_large}, Table~\ref{tab:sm-large}).

The first two observations also largely generalize across data geometries (Appendix~\ref{sm:toy}, Table~\ref{tab:sm-toy-summary}). On five synthetic datasets trained with the same schedule, persistence is $1.8$--$2.2$ times higher under DB than under UB in every case. The switching set is $26$--$46\%$ smaller under DB on Spiral, Checkerboard, Dartboard and Rings, and marginally smaller, within the seed-to-seed spread, on Blobs, which UB solves almost perfectly. Neither measure tracks the accuracy gain, however. The same smaller, persistent set appears where DB rescues the network, on Spiral and on Rings (test accuracy $79.0\pm6.7\%$ against $65.7\pm10.4\%$), and where it leaves accuracy unchanged within noise, on Checkerboard, Dartboard and Blobs.

Figure~\ref{fig:cost}d illustrates the functional side of this picture in one representative DB run. Late in training, accuracy is at ceiling for most of the period, drops in its middle part while the loss rises, and recovers within $5$--$10$ steps once the emphasis on the focal class has decreased, a small fraction of the $T=320$ steps of the period. Early in training, accuracy stays between $\approx0.35$ and $\approx0.5$ for almost the whole period. The abrupt onset and end of the dip ($t\approx116$ and $t\approx205$) coincide with the flooring of Eq.~\eqref{eq:collapse}: at these steps the number of examples of each non-focal class drops from $3$ to $2$ per batch, and later returns to $3$. The non-focal classes never vanish from the batch (at the peak, $g=A=20$, the batch holds $45$ focal-class examples and $2$ of each other class), but the step from $3$ to $2$ is a one-third cut in each non-focal class, which is why its effect on the loss is so abrupt. The dip therefore marks the interval in which the other classes are most under-represented, and the recovery is the functional counterpart of the switch that Fig.~\ref{fig:cost}a--c quantify in weight space. Fast recovery late in training is what one would expect from a switch carried by fewer, persistent weights, but it also coincides with the network having learned the task; our measurements show that the two views change together, not that one causes the other.

\section{A channel-narrowed CNN on CIFAR-10}
\label{sec:scaling}

The measurements above concern a single hidden layer trained on synthetic data. To check whether the accuracy effect of Sec.~\ref{sec:regime} carries over to a standard architecture and dataset, we apply the same batch-composition schedule [Eq.~\eqref{eq:ramp}] to a ResNet-10~\cite{he2016resnet} trained on CIFAR-10~\cite{krizhevsky2009cifar} ($C=10$). The network has four stages of one basic residual block each, with all channel widths scaled by a factor of $0.15$ to reduce capacity. It is trained for $100{,}000$ steps with batch size $128$ using SGD with momentum $0.9$ and weight decay $5\times10^{-4}$; the learning rate starts at $0.0679$ and is divided by $10$ after one and two thirds of training. We scan $14$ amplitudes from $A=1$ to $A=250$ and $15$ periods from $T=10$ to $T=1500$ steps (values listed in Appendix~\ref{sm:cnn}), with $5$ seeds per configuration. Accuracies are measured at the end of training on $5{,}000$-example subsets of the training and test sets. Under UB, averaged over the $15$ cells of the $A=1$ column, the network reaches $100.00\pm0.00\%$ training and $79.84\pm0.49\%$ test accuracy: unlike the small spiral network, it fits its training set. All values in this section are mean $\pm$ s.d.\ over seeds. In the heatmaps, the gain of each cell is measured relative to the UB cell ($A=1$) with the same period; the numbers quoted in the text instead use the mean of the whole $A=1$ column, which pools all UB runs. Both baselines give the same gain for the best configuration to within $0.01$ points (Appendix~\ref{sm:cnn}). We do not repeat the weight-level analysis of Sec.~\ref{sec:cost} here and leave it to future work.

\begin{figure*}[tb]
    \centering
    \includegraphics[width=\linewidth]{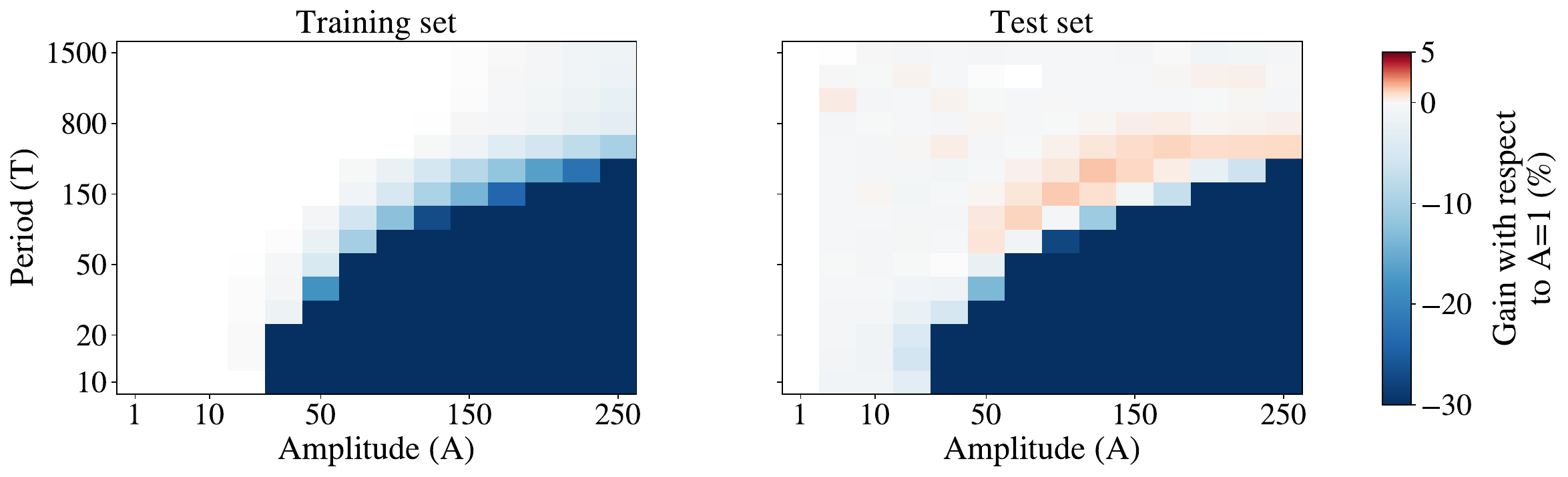}
    \caption{\textbf{Accuracy change on a channel-narrowed ResNet-10.} Gain in training (left) and test (right) accuracy, in percentage points, relative to the Uniform Batching (UB) run ($A=1$) with the same period $T$ (UB averaged over all periods: train $100.00\%$, test $79.84\%$), across amplitude $A$ and period $T$, for a ResNet-10 with channel widths scaled by $0.15$, trained on CIFAR-10. Each cell averages 5 seeds. The color scale is clipped at $-30$ points: dark blue cells are runs that collapse, mostly to chance level. Axes are categorical; only a subset of the tested values is labeled.}
    \label{fig:cnn-main}
\end{figure*}

A band of positive test-accuracy gain runs along the edge of the failure region, at large amplitudes and periods of one to a few hundred steps (Fig.~\ref{fig:cnn-main}): $29$ of the $195$ DB configurations exceed the UB test accuracy, $13$ of them by more than half a point. The maximum test accuracy is $81.27\pm0.38\%$ at $A=125$ and $T=200$, a gain of $+1.43$ points, about $2.3$ times the combined seed-to-seed s.d.\ of the two configurations ($0.62$), and neighboring cells along the same band reach comparable values [$81.01\%$ at $(A,T)=(75,100)$, $80.98\%$ at $(100,150)$ and $80.86\%$ at $(150,200)$]. Because this is the maximum over a grid selected on test accuracy, it is biased upward; the band of positive gain, rather than the single best cell, is the evidence for the effect. The gain comes with a \emph{decrease} in training accuracy, to $94.65\pm0.58\%$ for that configuration, and every configuration that improves test accuracy by more than half a point has training accuracy between $90\%$ and $99.5\%$, whereas configurations that still fit the training set gain at most $0.35$ points: here the oscillations act as a regularizer, trading training fit for generalization, rather than rescuing an optimization failure as the small network on the spiral. At large amplitudes combined with short periods, both training and test accuracy collapse, as in the failure region of the spiral, and the boundary of this region moves to longer periods as the amplitude grows. The size of the gain is comparable to that of the over-parametrized spiral network ($+1.74$ points), not to the rescue of the under-parametrized one ($+37.87$ points).

The architecture sweep of Appendix~\ref{sm:cnn} (Table~\ref{tab:sm-arch-summary}) indicates which models benefit. Alongside the narrow ResNet-10 we tested a plain CNN whose capacity is limited by a $50$-unit fully connected bottleneck, a VGG-style Myrtle5 network, and the same ResNet-10 at half width. The plain CNN shows the largest and broadest gain ($+2.00\pm0.96$ points at $A=75$, $T=1500$, with $42$ configurations improving test accuracy by more than one point), again with lower training accuracy; it does not quite fit its training set under UB ($98.30\%$). Myrtle5 and the half-width ResNet-10 gain at most about half a point, within noise. The benefit is therefore not restricted to networks that fail to fit their data, since the narrow ResNet-10 fits it; across the four architectures it appears in the two narrowest ones, which are also those with the lowest UB test accuracy. With two architectures on each side, we read this as consistent with, not as a demonstration of, a benefit concentrated in capacity-limited models.

\section{Discussion}

Our results show that Dynamical Batching (DB), relative to Uniform Batching (UB), changes both whether a capacity-limited network reaches a good solution and how the change between class-specialized states is organized in its weights. DB introduces no modular architecture, routing mechanism, task mask, or sparsity penalty; it only changes which class receives the strongest optimization pressure. Under DB, the change between class-emphasized states is carried by a set of weights that is about twice as persistent from cycle to cycle as under UB on every dataset we tested, and smaller on all of them.

This picture connects three lines of work. First, dynamic-objective methods show that changing the optimization target over time can improve trainability and generalization \cite{ruizgarcia2021}; we find that the same can be achieved through batch composition alone. Second, multitask-learning work identifies interference between task gradients as a central optimization obstacle \cite{sener2018multi,yu2020gradient,chen2018gradnorm}; our probe-gradient measurements show that per-class gradients nearly cancel under UB, equally when training plateaus and when it succeeds, so conflict alone does not explain the plateau, and that DB changes how the conflicting gradients are combined (Appendix~\ref{sm:gradients}). Third, modularity and continual-learning work shows that allocating different functions to distinguishable parameter subsets can reduce interference \cite{csordas2021neural,mallya2018piggyback,serra2018hat}, and optimization is known to concentrate in small subspaces or subnetworks \cite{gurari2018gradient,frankle2019lottery}. Our experiments suggest a bridge: temporal class cycling lets a network carry the class-to-class change in a stable subset of its parameters, without that subset being imposed.

The connection to nonequilibrium training is suggestive. Periodically varying related goals can produce modular organization without explicitly selecting for it \cite{kashtan2005}, can facilitate adaptation \cite{parter2008facilitated}, and has an optimal time scale \cite{sachdeva2020tuning}; \citet{falk2023} showed that periodic target switching can teach a physical material to change functionality through a small subset of design variables. We observe a related effect in gradient-trained networks: repeated exposure to different but related class objectives is associated with a small, persistent parameter footprint for switching. In our setting, however, this footprint is not a set of class-specific modules.

Several limitations are important. First, the weight-level evidence comes from one-hidden-layer MLPs on synthetic three-class tasks; on CIFAR-10 we show only an accuracy effect. Second, neither the size nor the persistence of the switching set tracks the accuracy gain. Third, our control is ordinary UB training, whose window-to-window motion is largely stochastic; a periodic emphasis on class-agnostic data subsets would separate effects of class structure from those of periodic forcing per se. Fourth, our notion of localization is weaker than the functional modularity of \citet{csordas2021neural} and \citet{hod2021modularity}, and our measurements are correlational: a causal test would freeze the switching set late in training and ask whether switching and recovery fail, compared with freezing the same number of random weights. Fifth, the protocol requires a meaningful partition of the data, and amplitude and period interact strongly; configurations were selected on test accuracy, and we have not compared DB with other regularizers or class-reweighting schemes on CIFAR-10.

Overall, the main result is not that cyclic training universally improves neural networks. Rather, it suggests that \emph{how optimization pressure is distributed in time can shape the internal organization of the solution}: a fixed architecture trained under a changing sequence of related sub-problems can develop a small, persistent parameter response to those sub-problems.

\section*{Data availability}

The protocol is fully specified in Sec.~\ref{sec:setup} [Eqs.~\eqref{eq:ramp}--\eqref{eq:collapse}], the weight-level measures in Sec.~\ref{sec:measures} and Appendix~\ref{sm:switch} [Eq.~\eqref{eq:oscillation}], and the probe-gradient test in Sec.~\ref{sec:regime} and Appendix~\ref{sm:gradients}. Dataset generation, grid-search configurations and baselines for the synthetic datasets and CNN architectures are reported in Appendixes~\ref{sm:toy} and \ref{sm:cnn}. The synthetic datasets are generated by the released code, and CIFAR-10 is publicly available~\cite{krizhevsky2009cifar}. Code and configuration files to reproduce all results are available
at \url{https://github.com/samuellozanoiglesias/dynamical_batching}
and archived on Zenodo~\cite{dbcode}.

\section*{Use of AI tools}

Generative AI tools were used to help refine hypotheses about the mechanism behind the observed accuracy gains, to provide feedback on experimental design (e.g., the choice of control conditions and ablations), to search for and summarize related literature while writing the Introduction and Discussion, to suggest structural edits to the manuscript, and to polish the writing. They were not used to generate datasets, to formulate or prove mathematical claims, to clean or reformat data, or to analyze data. The authors reviewed all AI-assisted content and take full responsibility for the content of this article.

\begin{acknowledgments}
The project that gave rise to these results received the support of a fellowship from Fundaci\'on Ram\'on Areces. We acknowledge support from Ram\'on y Cajal program (RYC2021-032055-I) funded by MCIN/AEI/10.13039/501100011033 and by European Union NextGenerationEU/PRTR, a Research Grant from HFSP (Ref.-No: RGEC33/2024) with the award DOI (\url{https://doi.org/10.52044/HFSP.RGEC332024.pc.gr.194170}) and grant PID2023-147067NB-I00 funded by MCIU/AEI/10.13039/501100011033 and by ERDF/EU. Samuel Lozano is a Graduate Fellow of the Madrid City Council (Spain) at the Residencia de Estudiantes, 2025-2026.
\end{acknowledgments}


\appendix
%
%
\section{Additional Toy Experiments}
\label{sm:toy}

Throughout the Appendices, DB denotes Dynamical Batching and UB Uniform Batching ($A=1$). This appendix expands the toy-network experiments of the main text. We first show the training curves behind the spiral results, then test robustness to batch size on the spiral, and finally evaluate the weight-level measures on four additional synthetic datasets. The purpose is not to claim universal optimal hyper-parameters, but to determine whether the qualitative pattern survives changes in data geometry.

\subsection{Training curves on the spiral: UB plateaus, DB does not}
\label{sm:curves}
\begin{figure*}[tb]
\centering
\includegraphics[width=\textwidth]{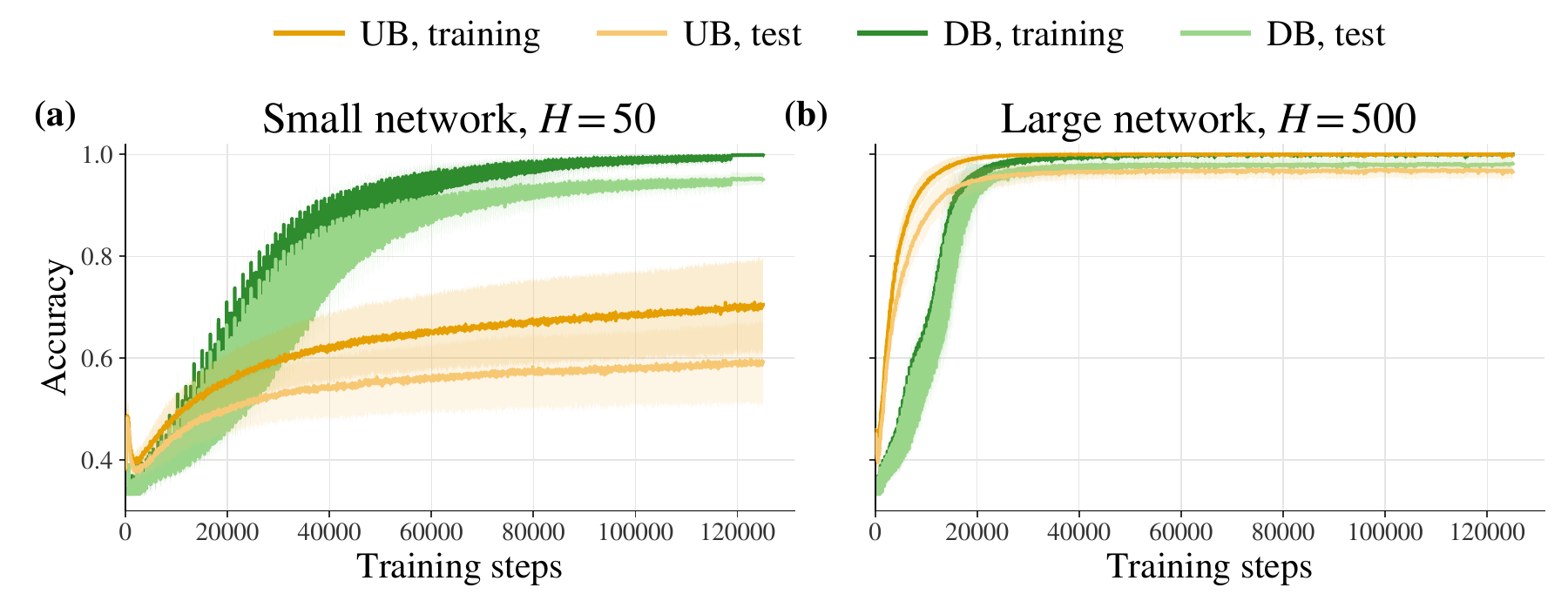}
\caption{\textbf{Training curves on the spiral.} Full-dataset training (dark) and test (light) accuracy over training for UB (amber) and DB (green); lines are the mean and bands $\pm1$ s.d.\ across 20 seeds. The final $5\%$ of training is uniform for both protocols. \textbf{(a)} Small network ($H=50$), $(A,T)=(20,320)$. After an initial rise, UB levels off, gaining less than $10$ points of training accuracy over the last two thirds of training, and ends at $70.5\pm8.9\%$ training and $59.2\pm7.7\%$ test accuracy. DB overtakes it after $\approx20{,}000$ steps and ends at $99.9\pm0.3\%$ and $95.1\pm1.3\%$. The fast oscillation of the DB curves follows the periodic class emphasis. \textbf{(b)} Large network ($H=500$), $(A,T)=(10,150)$. Both protocols fit the training set ($99.9\pm0.5\%$ UB, $100.0\pm0.0\%$ DB), UB faster than DB; test accuracy ends at $96.6\pm1.5\%$ (UB) and $98.2\pm1.2\%$ (DB).}
\label{fig:sm_accuracy}
\end{figure*}

Figure~\ref{fig:sm_accuracy} shows training and test accuracy over the whole run for the 20-seed experiments of Sec.~\ref{sec:cost} and Appendix~\ref{sm:large}. For the small network (panel a), UB rises during the first $\approx20{,}000$ steps and then improves only slowly: over the last two thirds of training its training accuracy gains less than $10$ points, ending at $70.5\pm8.9\%$ ($59.2\pm7.7\%$ test), with a wide spread across seeds. The same network under DB passes the UB curve after $\approx20{,}000$ steps and ends at $99.9\pm0.3\%$ training and $95.1\pm1.3\%$ test accuracy. For the large network (panel b), both protocols fit the training set, and UB does so faster than DB; DB ends with slightly higher test accuracy ($98.2\pm1.2\%$ against $96.6\pm1.5\%$).

UB accuracy in the small network is still increasing slowly at the end of training, so we tested whether a longer budget closes the gap. Doubling the training length to $250{,}000$ steps, with everything else unchanged (20 seeds per protocol, uniform final $5\%$), barely moves UB: training accuracy rises to $75.3\pm11.3\%$ and test accuracy to $62.4\pm10.8\%$, that is, by $4.8$ and $3.2$ points for $125{,}000$ additional steps. DB reaches $100.0\pm0.0\%$ training and $94.9\pm1.4\%$ test accuracy, so the gap remains $24.7$ points in training and $32.5$ points in test accuracy (Table~\ref{tab:sm-long}). The difference is therefore not a matter of DB reaching the UB solution sooner within a fixed budget. The weight-level picture is also unchanged at the longer budget: under DB the switching set is about half the size of that under UB and about $1.6$ times as persistent. All other small-network results use $125{,}000$ steps, for consistency across datasets and protocols.

\begin{table*}[tb]
\centering
\caption{Spiral, small network, $(A,T)=(20,320)$, trained for $250{,}000$ steps (twice the budget used elsewhere), mean $\pm$ s.d.\ over 20 seeds. Columns as in Table~\ref{tab:switching_final}.}
\label{tab:sm-long}
\begin{tabular}{lcccccc}
\toprule
Method & Train acc.\ (\%) & Test acc.\ (\%) & $|\Sset|/n$ & Persistence $J$ & $d$ \\
\midrule
UB & $75.3\pm11.3$ & $62.4\pm10.8$ & $0.29\pm0.08$ & $0.34\pm0.06$ & $0.18\pm0.13$ \\
DB & $100.0\pm0.0$ & $94.9\pm1.4$  & $0.15\pm0.07$ & $0.55\pm0.19$ & $0.38\pm0.17$ \\
\bottomrule
\end{tabular}%
\end{table*}

\subsection{Batch-size robustness on the spiral}

\begin{figure*}[tb]
\centering
\includegraphics[width=0.8\textwidth]{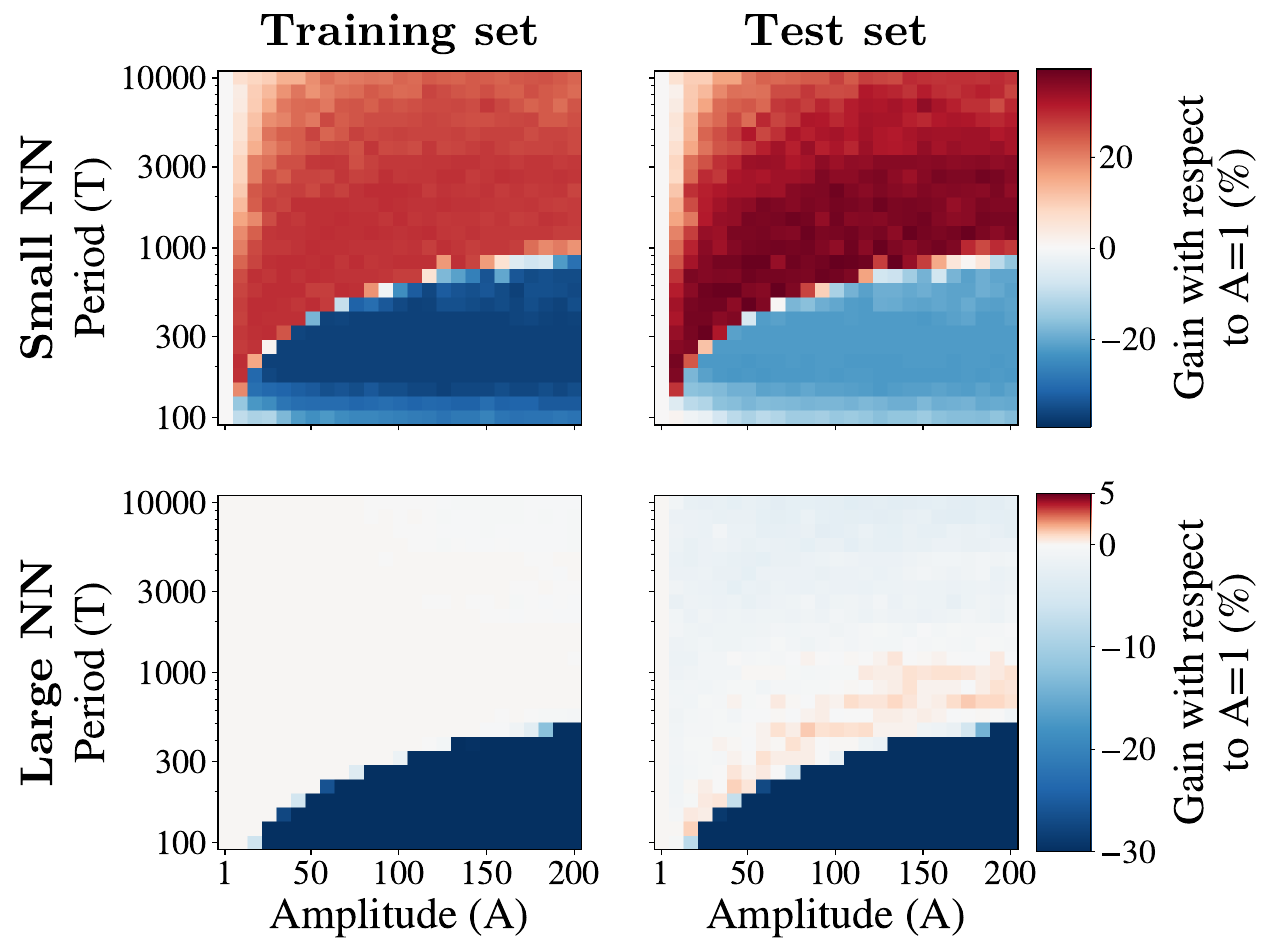}
\caption{\textbf{Grid search on the spiral at a larger batch size ($B=200$).} Accuracy gain in percentage points over the UB run ($A=1$) with the same period $T$, across oscillation amplitude $A$ and period $T$, for the small ($H=50$) and large ($H=500$) networks; training and test gain are shown separately. As for $B=50$ in the main text, a broad region of positive gain unlocks the capacity-constrained network, bordered by a failure region at extreme amplitudes and short periods where the emphasis shifts faster than the network can consolidate. Each cell represents the average of 5 independent seeds.}
\label{fig:sm-toy-grid}
\end{figure*}

We swept period $T$ and amplitude $A$ for two network widths ($H=50$ and $H=500$ hidden units) on the three-arm spiral, with the batch size increased from $B=50$ to $B=200$ and the optimizer and architectures unchanged. The corresponding grid maps (Fig.~\ref{fig:sm-toy-grid}, the analogue of Fig.~\ref{fig:phase}a without decision boundaries) show the same qualitative structure: a broad region of positive gain and a failure region at large amplitudes and short periods.

For the small network ($H=50$), the UB baseline reaches $70.20\%$ training and $55.60\%$ test accuracy, and the configuration at $(A,T)=(70,550)$ reaches $99.33\%$ training and $94.93\%$ test accuracy. For the large network ($H=500$), the UB baseline reaches $99.33\%$ training and $96.13\%$ test accuracy, while the best tested oscillating configuration reaches $97.33\%$ test accuracy at $(A, T) = (80,500)$. The numerical optima differ from those of the main-text grid because the batch size and sweep ranges differ; we therefore treat the grid search as evidence for a broad operating regime rather than for a universal optimal $(A,T)$.

Batch composition is deterministic (Eq.~\eqref{eq:collapse}), so the protocol contains no multinomial sampling noise at any batch size. What the batch size changes is the effect of the flooring: the realized composition, and the amplitude above which non-focal classes vanish from the batch ($g(t_T)>B-C+1$), both depend jointly on $A$, $B$ and $C$. That the phase structure survives a fourfold change in $B$ suggests that it is not tied to a particular flooring pattern. The failure region does move: at B=200 it recedes to somewhat shorter periods, most visibly for the large network, so the boundary depends on the batch size as well as on A and T.

\subsection{Weight-level measures across five toy datasets}

To test whether the pattern of Sec.~\ref{sec:cost} is specific to the spiral geometry, we repeated the measurements on five synthetic benchmarks. All use the small MLP  of the main text (one hidden layer, $50$ units, ReLU activations, Glorot initialization) trained for $125{,}000$ steps with $B=50$, with 20 seeds per configuration.

\begin{itemize}
    \item \textbf{Spiral:} 3 classes, 100 points per class. Class $j$ follows an arm with angular offset $2\pi j/3$: its points have radii evenly spaced from $0.01$ to $1$ and angles evenly spaced over an interval of length $4\pi$ (two full turns), and Gaussian noise of standard deviation $0.2$ is added to each angle.
    \item \textbf{Blobs:} 3 classes, 100 points per class. Fifteen Gaussian modes are equally spaced on a circle of radius $0.6$ and assigned to the three classes in rotation, so that each class is a mixture of 5 modes $72^\circ$ apart and neighboring modes belong to different classes; noise standard deviation $0.05$.
    \item \textbf{Checkerboard:} 3 classes, 100 points per class, drawn from the cells of a $7\times7$ grid whose labels alternate among the three classes; noise standard deviation $0.1$.
    \item \textbf{Dartboard:} 3 classes, 100 points per class. The disc between radii $0.1$ and $1.0$ is divided into 3 radial rings and 12 angular sectors, and the cell in ring $r$ and sector $s$ is assigned class $(r+s)\bmod 3$. Points are sampled uniformly in area within the central part of each cell, leaving an empty margin of $15\%$ of the cell's radial and angular width on each side; Gaussian noise of standard deviation $0.03$ is then added to both coordinates.
    \item \textbf{Rings:} 3 classes, 100 points per class. Each class consists of 2 concentric rings, interleaved with those of the other classes, starting at radius $0.1$ with a radial separation of $0.15$ between consecutive rings; noise standard deviation $0.05$.
\end{itemize}

\begin{figure*}[tb]
    \centering
    \includegraphics[width=\textwidth]{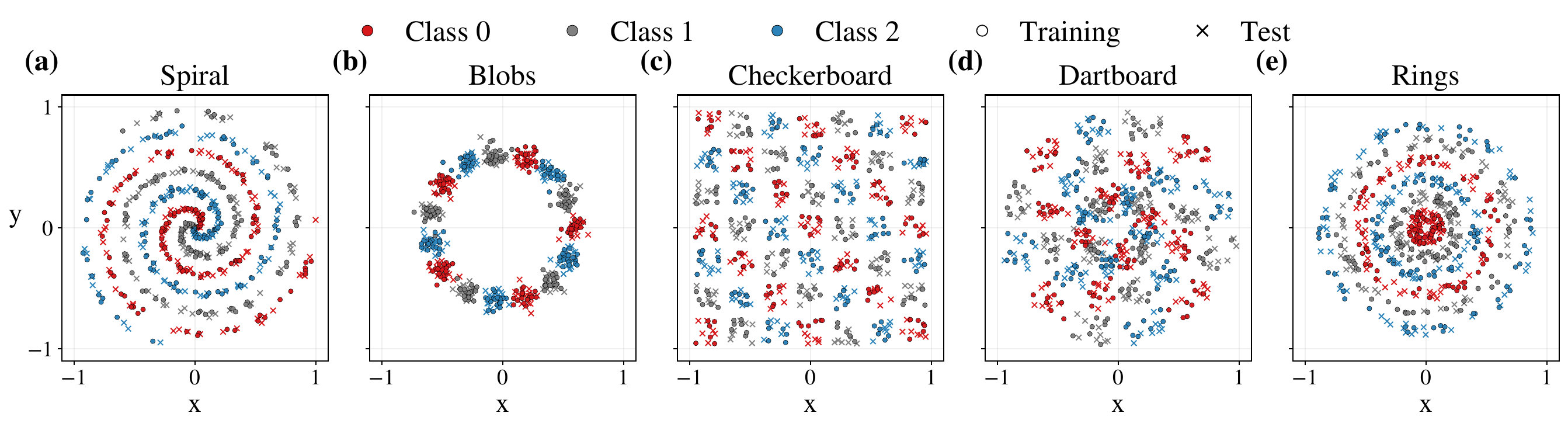}
    \caption{\textbf{The five synthetic datasets.} Training samples (circles) and test samples (crosses), colored by class, for \textbf{(a)} Spiral, \textbf{(b)} Blobs, \textbf{(c)} Checkerboard, \textbf{(d)} Dartboard and \textbf{(e)} Rings, each with 3 classes and 100 training and 100 test points per class, all within the region $[-1,1]\times[-1,1]$. Each panel shows one realization (seed $0$) generated from the exact configuration used for training; every training seed draws a new realization.}
    \label{fig:sm-datasets}
\end{figure*}

Figure~\ref{fig:sm-datasets} shows the five datasets. All five are normalized to the region $[-1,1]\times[-1,1]$, so the networks receive inputs on the same scale for every geometry. For every dataset, the test set ($100$ points per class) is generated by the same procedure as the training set with a different random seed. Each training seed also sets the dataset seed, so each seed draws a new dataset (training and test sets) and a new initialization, and the spread across seeds reported throughout includes dataset-to-dataset variability. For the spiral, whose radii are evenly spaced, training and test points therefore lie at the same radii and differ only in their angular noise.

For all datasets we use $(A,T)=(20,320)$, the values selected for the main-text spiral experiment. This is a transfer test rather than a per-dataset optimization, so Table~\ref{tab:sm-toy-summary} should not be read as a comparison of the maximum attainable accuracy on each dataset.

\begin{table*}[tb]
\centering
\caption{Accuracy and weight-level measures across five toy datasets, $(A,T)=(20,320)$, mean $\pm$ s.d.\ over 20 seeds. Accuracies are measured at the end of training (after the uniform final phase); switching measures at the last cycle of each protocol's regime. $|\Sset|/n$: switching-set size; $J$: persistence of the switching set; $d$: switching distance (Sec.~\ref{sec:measures}).}
\label{tab:sm-toy-summary}
\begin{tabular}{llccccc}
\toprule
Dataset & Method & Train acc.\ (\%) & Test acc.\ (\%) & $|\Sset|/n$ & $J$ & $d$ \\
\midrule
Spiral       & UB & $70.5\pm8.9$  & $59.2\pm7.7$  & $0.32\pm0.04$ & $0.33\pm0.06$ & $0.24\pm0.19$ \\
             & DB & $99.9\pm0.3$  & $95.1\pm1.3$  & $0.19\pm0.06$ & $0.59\pm0.13$ & $0.64\pm0.14$ \\
\midrule
Blobs        & UB & $99.8\pm0.4$  & $97.8\pm1.1$  & $0.20\pm0.08$ & $0.35\pm0.10$ & $0.19\pm0.06$ \\
             & DB & $99.8\pm0.4$  & $96.7\pm1.6$  & $0.16\pm0.10$ & $0.71\pm0.10$ & $0.35\pm0.08$ \\
\midrule
Checkerboard & UB & $97.8\pm2.7$  & $89.7\pm4.1$  & $0.26\pm0.07$ & $0.33\pm0.06$ & $0.17\pm0.09$ \\
             & DB & $97.6\pm1.4$  & $90.9\pm1.7$  & $0.16\pm0.08$ & $0.73\pm0.14$ & $0.53\pm0.12$ \\
\midrule
Dartboard    & UB & $96.8\pm1.5$  & $83.8\pm3.0$  & $0.31\pm0.03$ & $0.34\pm0.06$ & $0.18\pm0.03$ \\
             & DB & $97.3\pm1.4$  & $85.3\pm2.3$  & $0.23\pm0.04$ & $0.61\pm0.05$ & $0.44\pm0.06$ \\
\midrule
Rings        & UB & $80.4\pm9.7$  & $65.7\pm10.4$ & $0.24\pm0.08$ & $0.31\pm0.07$ & $0.33\pm0.17$ \\
             & DB & $94.2\pm4.2$  & $79.0\pm6.7$  & $0.13\pm0.03$ & $0.63\pm0.05$ & $0.47\pm0.07$ \\
\bottomrule
\end{tabular}
\end{table*}

Three observations follow from Table~\ref{tab:sm-toy-summary}.

\paragraph{The switching set is more persistent under DB on every dataset.} On all five datasets the switching set is more persistent under DB than under UB, by a factor of $1.8$--$2.2$.

\paragraph{The switching set is smaller under DB on every dataset.} DB needs fewer weights than UB to carry $80\%$ of the class-to-class variation on Rings ($0.13$ vs.\ $0.24$), Spiral ($0.19$ vs.\ $0.32$), Checkerboard ($0.16$ vs.\ $0.26$) and Dartboard ($0.23$ vs.\ $0.31$), a reduction of $26$--$46\%$, and marginally fewer on Blobs ($0.16$ vs.\ $0.20$, within the seed-to-seed spread). The concentration reported for the spiral in Sec.~\ref{sec:cost} therefore transfers to all five geometries, although its magnitude varies.

\paragraph{Neither measure tracks the accuracy gain.} DB improves test accuracy substantially on Spiral ($+35.9$ points) and Rings ($+13.3$), and leaves it unchanged within seed-to-seed variability on Checkerboard ($+1.2$), Dartboard ($+1.5$) and Blobs ($-1.1$). Persistence is higher under DB on every dataset, and the switching set is smaller, so both properties appear with a gain (Spiral, Rings) and without one (Checkerboard, Dartboard, and Blobs). They are therefore a signature of the protocol, not of its accuracy benefit. The switching distance is larger under DB on every dataset, as expected, since only DB moves between class-specialized solutions (Sec.~\ref{sec:measures}).

Taken together, the five datasets show that under DB the switch is carried by largely the same weights from cycle to cycle on all five geometries, and by fewer weights than under UB (within noise on Blobs), whereas the accuracy gain is specific to some of them. Nor is this organization a by-product of an accuracy gap: the set is smaller and more persistent under DB on Checkerboard and Dartboard, where both protocols reach the same accuracy, just as on Spiral and Rings; it is also about twice as persistent on Blobs, which UB already solves; and it holds in the large network of Appendix~\ref{sm:large}, where both protocols fit the data.

\section{Weight-level switching analysis}
\label{sm:switch}

Unless stated otherwise, results use the setting of Sec.~\ref{sec:cost}: small network ($H=50$), $(A,T)=(20,320)$, 20 seeds per protocol, dynamical cycles only. The large network ($H=500$) uses $(A,T)=(10,150)$, the best configuration of its grid in Fig.~\ref{fig:phase}a, also with 20 seeds per protocol ($277$ cycles, of which the first $264$ are dynamical). Solid green is DB and dashed amber is UB; bands are $\pm1$ s.d.\ across seeds.

\subsection{Definition of the switching set}
Let $q_i\in\{0,\dots,C-1\}$ be the position of class $i$ in the cycle, $m_k(c)=\frac{1}{C}\sum_i\theta_{i,k}(c)$ the cycle mean of weight $k$ (since each $\theta_{i,k}(c)$ is itself a time average, $m_k(c)$ is the average of weight $k$ over the $CT/2$ central steps of the cycle), and $D_k(c)=m_k(c)-m_k(c-1)$ its drift between consecutive cycles. Because the $C$ class windows are consecutive in time, a weight that drifts by $D_k$ per cycle moves by $D_k/C$ from one class window to the next, and would differ between class states even without any class-specific change. For each weight we therefore remove a linear trend across the cycle, with its slope set by this drift, and quantify class-to-class switching by the variance of the residuals:
\begin{align}
    u_{i,k}(c)=\theta_{i,k}(c)-\alpha_k(c)-\beta_k(c)\,q_i,
    \qquad \beta_k(c)=\frac{D_k(c)}{C}, \\
    \alpha_k(c)=m_k(c)-\beta_k(c)\,\frac{C-1}{2},
    \qquad a_k^2(c)=\frac{1}{C}\sum_i u_{i,k}(c)^2 .
    \label{eq:oscillation}
\end{align}
    
The intercept centers the residuals, $\sum_i u_{i,k}(c)=0$, so $a_k^2(c)$ is the variance across classes of the de-trended states; equivalently, $u_{i,k}(c)=\theta_{i,k}(c)-m_k(c)-D_k(c)\,(q_i-(C-1)/2)/C$. In the first cycle of each regime there is no previous cycle, and no trend is removed. All quantities are computed from the stored class states $\btheta_i(c)$ of every cycle. The switching set $\Sset(c)$ is the smallest set of weights with $\sum_{k\in\Sset(c)}a_k^2(c)\ge0.8\sum_k a_k^2(c)$, obtained by ranking weights by $a_k^2(c)$.

\paragraph{Why the slope is taken from the previous cycle.} Two simpler choices fail for identifiable reasons. Without any trend removal, $u_{i,k}=\theta_{i,k}-m_k$, pure drift produces $u_{i,k}=D_k\,(q_i-(C-1)/2)/C$: a class pattern that repeats identically every cycle. A network that only drifts would then show a switching set made of its fastest-drifting weights, and that set would appear highly persistent. Fitting the slope by least squares to the $C$ class states of the current cycle removes this drift, but it also removes two of the $C$ degrees of freedom, the mean and the slope, leaving only $C-2$ residual directions. For $C=3$ a single direction is left: with classes indexed by their position in the cycle, $u_{\cdot,k}\propto(1,-2,1)$ and $a_k^2=(\theta_{0,k}-2\theta_{1,k}+\theta_{2,k})^2/18$. A weight that separates the first and the last class of the cycle, pattern $(+1,0,-1)$, would then have $a_k=0$ and never enter the switching set, and the result would depend on which class the cycle happens to start with. Within one cycle, a class pattern that is linear in cycle position cannot be distinguished from drift. The drift $D_k(c)$ is estimated from outside the cycle, from the previous cycle mean, and is zero under a closed limit cycle whatever the oscillation amplitude, which is why Eq.~\eqref{eq:oscillation} separates the two.

\begin{figure*}[tb]
\centering
\includegraphics[width=\textwidth]{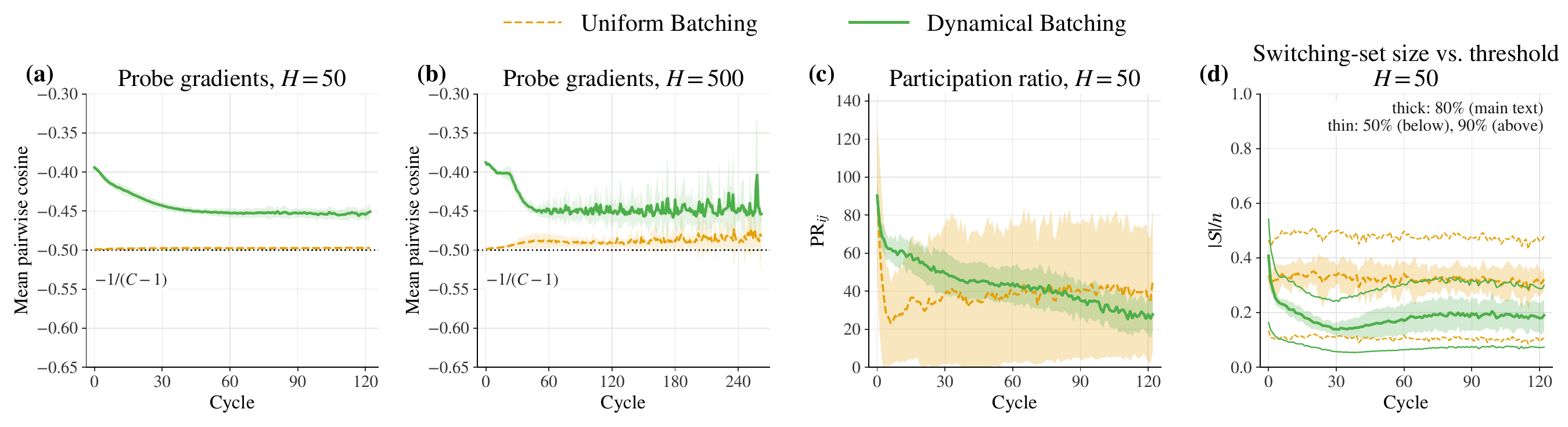}
\caption{\textbf{Additional measures on the spiral.} Solid green: DB; dashed amber: UB; bands $\pm1$ s.d.\ across 20 seeds. \textbf{(a)} Mean pairwise cosine between the per-class probe gradients $\nabla\mathcal{L}_k$, each computed on the training examples of a single class, with all three gradients evaluated at the same parameters (Appendix~\ref{sm:gradients}), small network ($H=50$). Under UB it sits at $-1/(C-1)=-0.5$ (dotted) throughout training, while training plateaus near $71\%$ (Fig.~\ref{fig:sm_accuracy}a). Under DB it starts at $\approx-0.39$ and settles at $\approx-0.45$ after $\approx40$ cycles: the class gradients remain strongly opposed but do not cancel. \textbf{(b)} The same for the large network ($H=500$). UB again sits close to $-0.5$ (between $-0.50$ and $-0.48$), although it fits the training set ($\geq99.8\%$): the arrangement is generic under UB, not a signature of failure. DB stays near $-0.40$ for the first $\approx25$ cycles and settles at $\approx-0.45$ after $\approx50$ cycles. \textbf{(c)} Participation ratio $\mathrm{PR}_{ij}=(\sum_k\Delta_k^2)^2/\sum_k\Delta_k^4$ of $\Delta=\btheta_i-\btheta_j$ \citep{ingrosso2022}, the effective number of weights carrying the displacement, small network. Under DB it falls from $90$ to $28$, agreeing in direction with $|\Sset|/n$ (Fig.~\ref{fig:cost}a); under UB it drops within the first cycles and then rises to $40$, with large seed-to-seed variability, so the two protocols separate only after cycle $75$--$80$. \textbf{(d)} Switching-set size $|\Sset|/n$ under DB (solid green) and UB (dashed amber) for energy thresholds of $50\%$ (thin, lower), $80\%$ (thick, main text) and $90\%$ (thin, upper). After the first few cycles, each DB curve lies below the UB curve of the same threshold, ending at $0.07$, $0.19$ and $0.30$ against $0.10$, $0.32$ and $0.47$: DB needs fewer weights than UB to carry the same fraction of the class-to-class variation, whichever fraction is chosen.}
\label{fig:sm_spiral}\label{fig:sm_gradients}
\end{figure*}

\subsection{Per-class gradient geometry}
\label{sm:gradients}

\paragraph{What is measured.} For each class $k$, $\nabla\mathcal{L}_k(\btheta)$ is the gradient, with respect to all network parameters, of the mean cross-entropy over a probe set that contains only examples of class $k$. The probe sets are drawn once, before training, and kept fixed throughout; with $100$ training examples per class, each contains all training examples of its class. The gradient is a diagnostic computed on stored parameters, not the gradient of any training mini-batch. This distinction matters for UB. A UB mini-batch mixes the three classes, and its gradient is approximately the balanced sum $\frac{1}{C}\sum_k\nabla\mathcal{L}_k$; comparing such mixed gradients with each other would say nothing about conflict between classes. The probe gradients instead isolate the contribution of each class, for UB and DB alike.

In each cycle, and at each of the $C$ class states $\btheta_i(c)$ used for the switching measures (Sec.~\ref{sec:measures}), we evaluate all $C$ probe gradients $\nabla\mathcal{L}_k(\btheta_i(c))$ at that same point and take their mean pairwise cosine; the reported value is the mean over the $C$ states of the cycle. Evaluating the $C$ gradients at the same parameters is what the argument below requires. Evaluating each class gradient at its own window center instead compares gradients at different parameters. Under DB, whose class states are far apart early in training, this produces an apparent cosine near $0$ in the first cycles that disappears when the gradients are evaluated at the same point.

\paragraph{Why $-1/(C-1)$.} At a stationary point of the balanced objective, $\sum_k\nabla\mathcal{L}_k=0$. Taking the inner product with $\nabla\mathcal{L}_j$ gives $\lVert\nabla\mathcal{L}_j\rVert^2=-\sum_{k\neq j}\langle\nabla\mathcal{L}_k,\nabla\mathcal{L}_j\rangle$, so for equal norms the mean pairwise cosine is exactly $-1/(C-1)$. A cosine of $-0.5$ therefore indicates that the per-class gradients nearly cancel, so that the balanced step is a small residual of large, opposing class gradients. One might suspect that this cancellation is confined to the output layer: given the current hidden features, the output layer solves a convex problem that balanced sampling brings quickly to its optimum, where its per-class gradients cancel whatever the quality of the features. Our measurements do not support this: restricted to the hidden-layer parameters or to the readout parameters alone, the cosine takes the same values as over all parameters, in both networks, close to $-0.5$ under UB and $\approx-0.45$ under DB.
	
The converse also deserves care. A mean cosine of $-1/(C-1)$ implies exact cancellation only for gradients of equal norm; the cosine alone is therefore evidence of near-cancellation, not a proof of it. Moreover, at a class-emphasized DB state the optimizer follows the weighted sum $\sum_k p_k\nabla\mathcal{L}_k$, so near-stationarity of that weighted objective does not require the unweighted sum to vanish; a cosine above $-0.5$ under DB is therefore expected and should not be read as reduced conflict.

\paragraph{Results.} Under UB (Fig.~\ref{fig:sm_spiral}a,b), the per-class gradients cancel throughout training in both networks, whether training ultimately succeeds or not. The cancelling arrangement is therefore the generic state of balanced training at this scale. It indicates that UB training stays close to a stationary point of the balanced loss over the time scale of a cycle, a poor one in the small network and a good one in the large network, and is not by itself the signature of a failed run. What distinguishes the small network is that it stays at a poor solution while in this arrangement (Sec.~\ref{sec:regime}, Fig.~\ref{fig:sm_accuracy}a). Under DB, the per-class gradients at the class-emphasized states are almost as strongly opposed ($\approx-0.39$ early in training, $\approx-0.45$ after $\approx40$--$50$ cycles in both networks) but do not cancel. DB therefore does not remove the conflict between classes. What it changes is the combination of the conflicting gradients that the optimizer follows: at each point, the batch weights the focal class more heavily, so the step is not the near-cancelling balanced sum. Whether this is what lets DB leave the poor solution remains a hypothesis.

\subsection{Robustness of the concentration measure}

The participation ratio, a threshold-free measure of how many weights carry the displacement between two class states, agrees in direction with the switching-set size, but separates the two protocols only late in training (Fig.~\ref{fig:sm_spiral}c). The switching-set size depends on the energy threshold, but its ordering between the protocols does not: at $50\%$, $80\%$ and $90\%$ it ends at $0.07$, $0.19$ and $0.30$ under DB against $0.10$, $0.32$ and $0.47$ under UB (Fig.~\ref{fig:sm_spiral}d).

\subsection{Persistence does not require an accuracy gap: the large network}
\label{sm:large}

In the small network, DB fits the data and UB does not, so differences in weight dynamics could in principle be a by-product of the accuracy gap. The large network removes this confound: both protocols fit the training set (Fig.~\ref{fig:sm_accuracy}b, Table~\ref{tab:sm-large}): training accuracy $99.9\pm0.5\%$ under UB and $100.0\pm0.0\%$ under DB, test accuracy $96.6\pm1.5\%$ and $98.2\pm1.2\%$, respectively (20 seeds).

\begin{table*}[tb]
\centering
\caption{Spiral, large network ($H=500$), $(A,T)=(10,150)$, mean $\pm$ s.d.\ over 20 seeds. Columns as in Table~\ref{tab:switching_final}.}
\label{tab:sm-large}
\begin{tabular}{lccccc}
\toprule
Method & Train acc.\ (\%) & Test acc.\ (\%) & $|\Sset|/n$ & Persistence $J$ & $d$ \\
\midrule
UB & $99.9\pm0.5$  & $96.6\pm1.5$ & $0.15\pm0.06$ & $0.34\pm0.14$ & $0.13\pm0.11$ \\
DB & $100.0\pm0.0$ & $98.2\pm1.2$ & $0.03\pm0.03$ & $0.70\pm0.20$ & $0.18\pm0.12$ \\
\bottomrule
\end{tabular}%
\end{table*}

\begin{figure*}[tb]
\centering
\includegraphics[width=\textwidth]{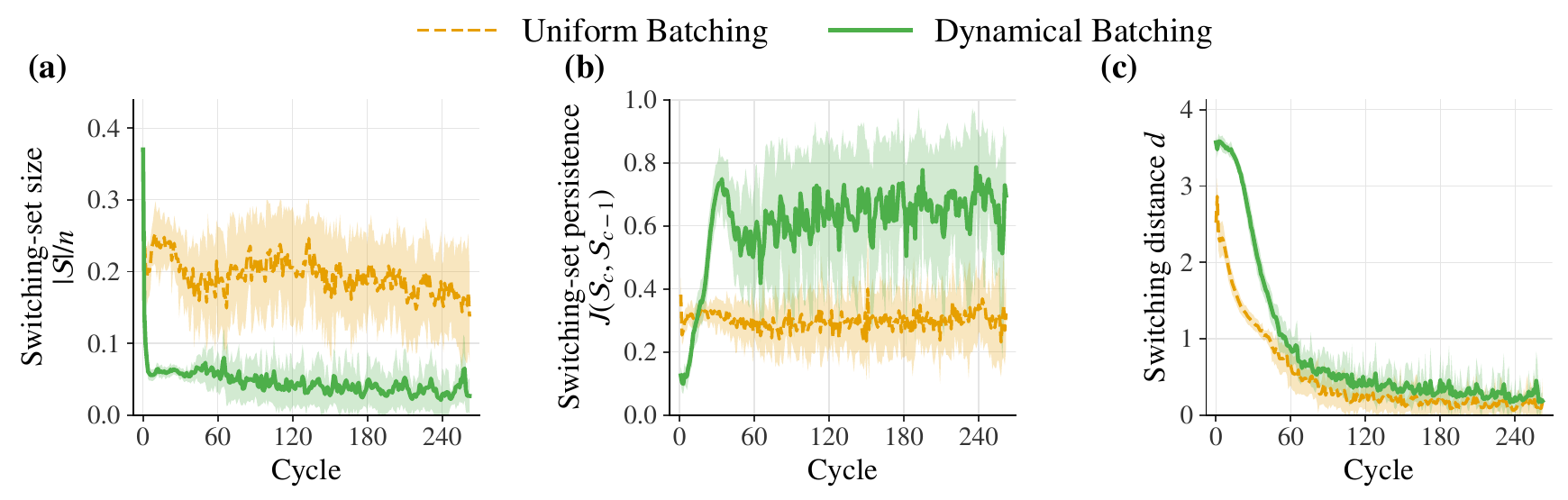}
\caption{\textbf{Studying the structure of the network (analogous to Fig.~\ref{fig:cost}a--c) for the large network ($H=500$), where both protocols fit the training set.} $(A,T)=(10,150)$, giving $277$ cycles, of which $264$ are dynamical. \textbf{(a)} Switching-set size $|\Sset|/n$: DB drops to $0.06$ within a few cycles and ends at $0.03$; UB stays at $0.15$--$0.25$ and ends at $0.15$. The concentration is stronger than in the small network (Fig.~\ref{fig:cost}a). \textbf{(b)} Persistence of the switching set (Jaccard overlap of consecutive cycles): DB starts below UB ($0.1$) while the switching set is being formed, rises to $0.7$ by cycle $\approx35$ and stays at $0.55$--$0.75$; UB stays at $0.30$--$0.34$. \textbf{(c)} Switching distance: DB falls from $3.6$ to $0.18$, about twentyfold, closely tracking the drift of ordinary training over the same windows (UB, from $2.8$ to $0.13$).}
\label{fig:sm_large}
\end{figure*}

The first two main-text observations hold, and are stronger, when there is no accuracy gap to explain them (Fig.~\ref{fig:sm_large}a,b). The third needs the same qualification as in the main text: the distance between class states falls, but so does the drift of ordinary UB training over the same windows (from $2.8$ to $0.13$), and the two end close to each other (Fig.~\ref{fig:sm_large}c).

\section{Additional CNN architectures on CIFAR-10}
\label{sm:cnn}

We repeated the grid search of Sec.~\ref{sec:scaling} on three further architectures to determine which models benefit from the protocol: a plain CNN with a fully connected bottleneck, a VGG-style Myrtle5 network, and the ResNet-10 of the main text at half width.

\begin{figure*}[tb]
\centering
\newcommand{\gridrow}[2]{\parbox[c]{2.2em}{\textbf{(#1)}}\parbox[c]{0.8\textwidth}{\includegraphics[width=0.8\textwidth]{#2}}\\[3pt]}
\gridrow{a}{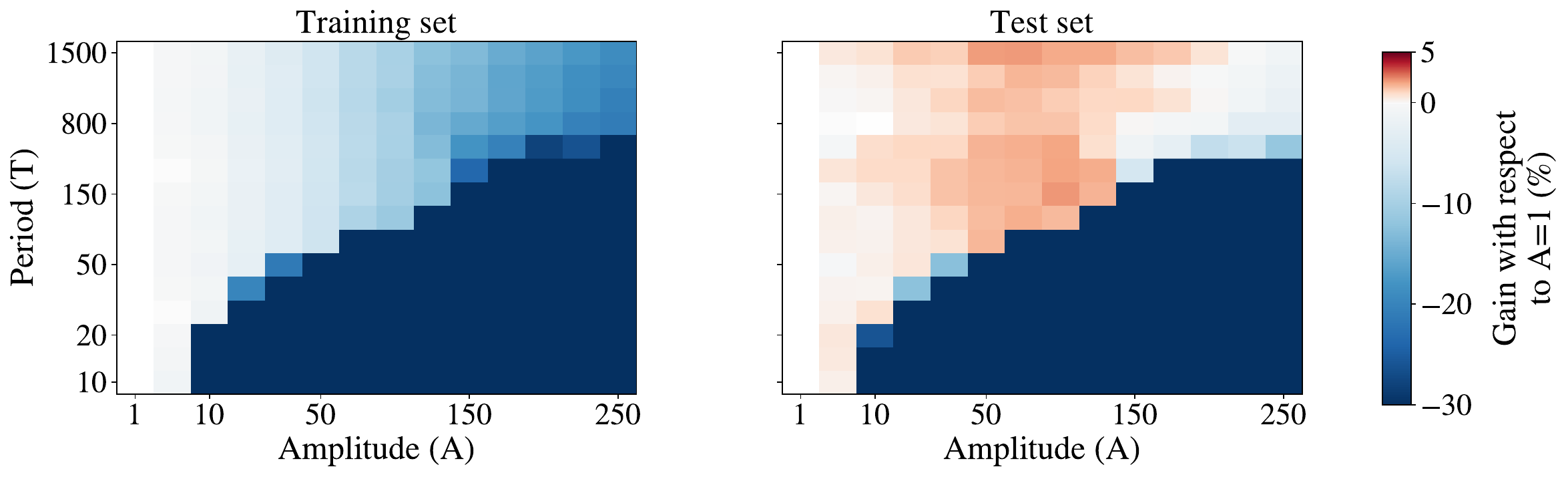}
\gridrow{b}{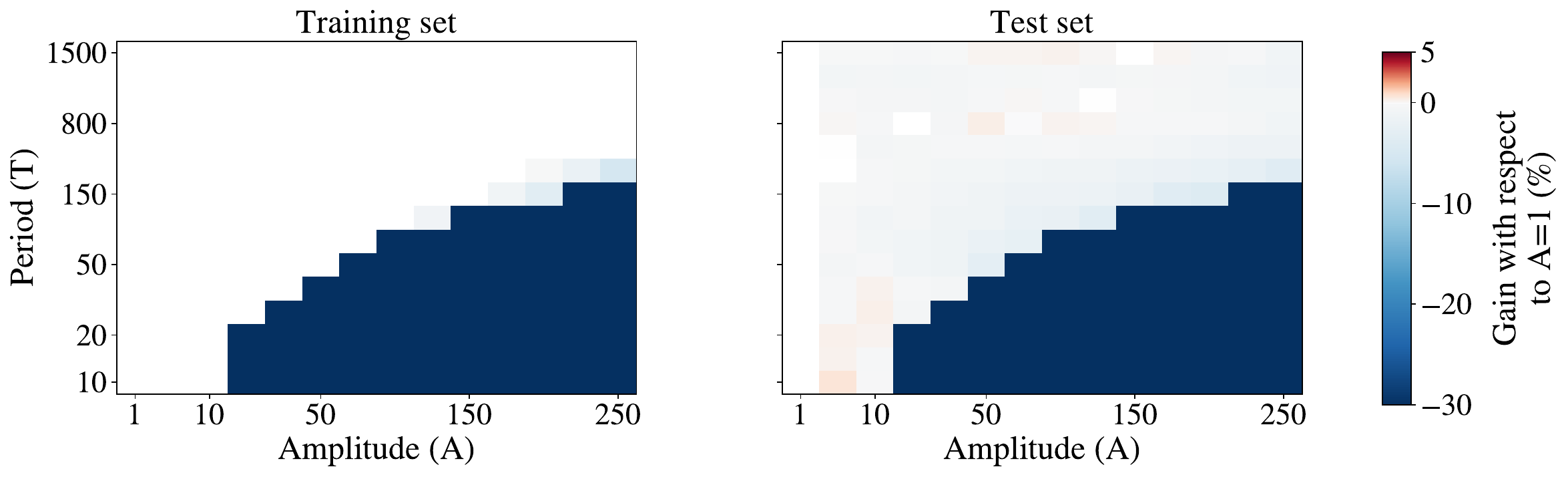}
\gridrow{c}{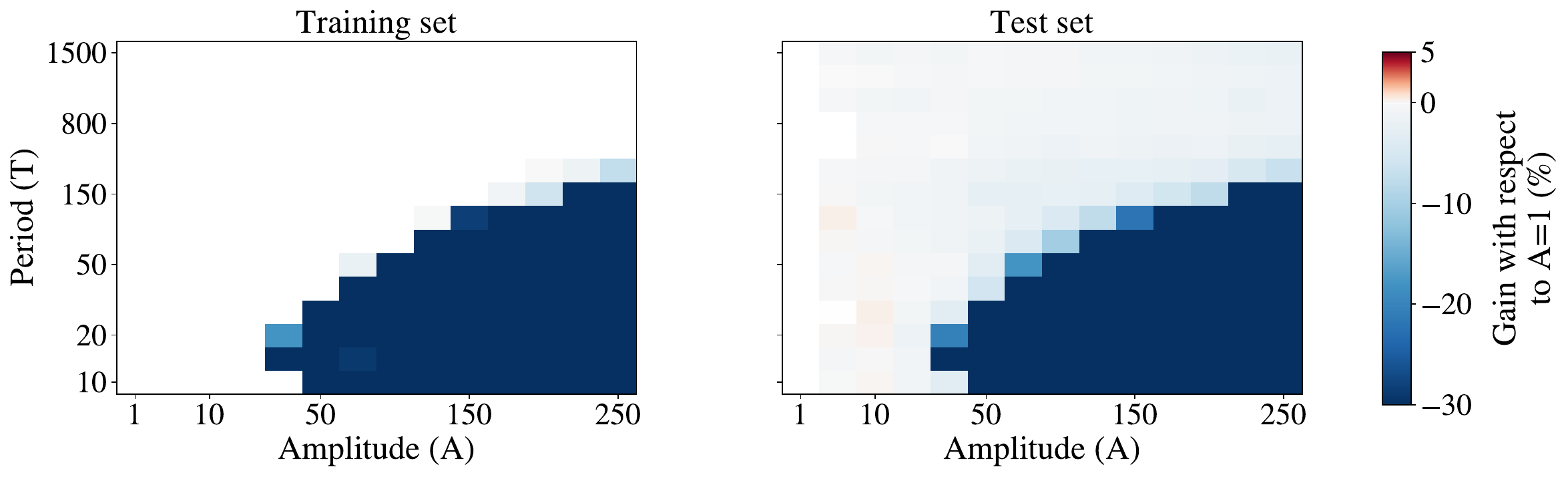}
\caption{\textbf{Grid searches for the additional CNN architectures.} Gain in training (left) and test (right) accuracy, in percentage points, relative to the UB run ($A=1$) with the same period $T$, across amplitude $A$ and period $T$, on CIFAR-10. Each cell averages $5$ seeds. The color scale is clipped at $-30$ points; dark blue cells are runs that collapse. Axes are categorical, as in Fig.~\ref{fig:cnn-main}. \textbf{(a)} Plain CNN with a $50$-unit fully connected bottleneck (UB: train $98.30\%$, test $75.97\%$): a broad region of positive test gain at intermediate to large amplitudes and periods of about a hundred steps or more, with reduced training accuracy. \textbf{(b)} Myrtle5 (UB: $100.00\%$, $85.10\%$). \textbf{(c)} ResNet-10 at width $0.5$ (UB: $100.00\%$, $85.98\%$). In (b) and (c) the test accuracy outside the failure region is flat to within seed-to-seed noise, and the training set is fitted wherever training does not collapse.}
\label{fig:sm-cnn-grids}
\end{figure*}

All architectures are trained for $100{,}000$ steps with batch size $128$ using SGD with momentum $0.9$ and weight decay $5\times10^{-4}$, without gradient clipping or Nesterov momentum. The initial learning rate is $0.05$ for the plain CNN and $0.0679$ for all other architectures, and in every case it is divided by $10$ at steps $33{,}333$ and $66{,}666$, one and two thirds of training. Every grid covers the same $14$ amplitudes, $A\in\{1,\allowbreak 5,\allowbreak 10,\allowbreak 20,\allowbreak 30,\allowbreak 50,\allowbreak 75,\allowbreak 100,\allowbreak 125,\allowbreak 150,\allowbreak 175,\allowbreak 200,\allowbreak 225,\allowbreak 250\}$, and $15$ periods, $T\in\{10,\allowbreak 15,\allowbreak 20,\allowbreak 30,\allowbreak 40,\allowbreak 50,\allowbreak 75,\allowbreak 100,\allowbreak 150,\allowbreak 200,\allowbreak 400,\allowbreak 800,\allowbreak 1000,\allowbreak 1200,\allowbreak 1500\}$ steps, that is, $210$ configurations of which $195$ are DB ($A>1$). Training and test accuracy are evaluated at the end of training on $5{,}000$-example subsets of the training and test sets. The UB baseline is the mean over the $15$ cells of the $A=1$ column, whose runs differ only by seed-to-seed and run-to-run variability, and its uncertainty is the s.d.\ over all runs in that column. In the heatmaps (Figs.~\ref{fig:cnn-main} and \ref{fig:sm-cnn-grids}), the gain of each cell is instead computed relative to the $A=1$ cell with the same period $T$. The two baselines differ only by the variability between $A=1$ cells (UB test accuracy $79.61$--$80.16\%$ across periods for the narrow ResNet-10), and the best-cell gains they give differ by at most $0.32$ points: $+1.42$ against $+1.43$ for the narrow ResNet-10, $+2.15$ against $+2.00$ for the plain CNN, $+0.21$ against $+0.53$ for Myrtle5 and $+0.08$ against $+0.22$ for the half-width ResNet-10 (same-$T$ against pooled baseline). No conclusion depends on this choice. Uncertainties of single configurations are s.d.\ across seeds; for a difference we quote the combined s.d.\ $\sqrt{s_{\mathrm{base}}^2+s_{\mathrm{best}}^2}$. All ``best'' values are maxima over the grid and are therefore biased upward by the selection, equally for every architecture. All heatmap's points include the average over 5 independent seeds.

A test-accuracy gain beyond the seed-to-seed variability appears only in the two narrowest models, the ResNet-10 at width $0.15$ (main text) and the plain CNN with a $50$-unit bottleneck, and in both it comes with lower training accuracy. The two wider models, Myrtle5 and the ResNet-10 at width $0.5$, show no gain beyond noise. Whether a model fits its training set under UB does not separate the two groups: the narrow ResNet-10 reaches $100\%$ training accuracy under UB and still benefits. With two architectures on each side, the protocol appears useful primarily in capacity-limited networks, but we have not isolated whether width, the resulting test accuracy, or another property of these models is the relevant variable, nor the mechanism in these CNNs. In every architecture, large amplitudes combined with short periods make training collapse, often to chance level, and the failure boundary moves to longer periods as the amplitude grows (Figs.~\ref{fig:cnn-main} and \ref{fig:sm-cnn-grids}).

\begin{table*}[tb]
\centering
\caption{Architectures tested on CIFAR-10 and their response to DB. Test-accuracy gains beyond seed-to-seed noise appear only in the two capacity-limited architectures (top two rows). ``Base'' values are UB ($A=1$) means over the $15$ cells of the $A=1$ column; ``Best test'' is the maximum over the DB configurations ($A>1$) of the grid, and ``Train at best'' the training accuracy of that same configuration. Every configuration is run with $5$ seeds. Values are mean $\pm$ s.d.\ over seeds (accuracies in \%); the test gain is given with the combined s.d.\ $\sqrt{s_{\mathrm{base}}^2+s_{\mathrm{best}}^2}$.}
\label{tab:sm-arch-summary}
\resizebox{\textwidth}{!}{%
\begin{tabular}{lccccccc}
\toprule
Architecture & Fits train (UB)? & Base train & Base test & Best test & $(A,T)$ & Train at best & Test gain \\
\midrule
ResNet-10 (width 0.15) (main text) & Yes & $100.00\pm0.00$ & $79.84\pm0.49$ & $81.27\pm0.38$ & $(125,200)$ & $94.65\pm0.58$ & Yes ($+1.43\pm0.62$) \\
Plain CNN + FC bottleneck (50) & No & $98.30\pm0.20$ & $75.97\pm0.75$ & $77.96\pm0.60$ & $(75,1500)$ & $90.07\pm0.35$ & Yes ($+2.00\pm0.96$) \\
\midrule
Myrtle5 (base width 64) & Yes & $100.00\pm0.00$ & $85.10\pm0.47$ & $85.64\pm0.59$ & $(5,15)$ & $100.00\pm0.00$ & No ($+0.53\pm0.75$) \\
ResNet-10 (width 0.5) & Yes & $100.00\pm0.00$ & $85.98\pm0.48$ & $86.20\pm0.07$ & $(10,50)$ & $100.00\pm0.00$ & No ($+0.22\pm0.48$) \\
\bottomrule
\end{tabular}%
}
\end{table*}
 
\subsection{Plain CNN with a fully connected bottleneck}
 
Three convolutional stages with $16$, $32$ and $64$ channels, $3\times3$ kernels, ReLU activations, no batch normalization, max pooling after every stage and no residual connections, followed by a $50$-unit fully connected bottleneck before the $10$-way classifier. Its capacity is limited by the bottleneck, like that of the small spiral network by its hidden layer.
 
\textbf{Results:} The UB baseline reaches $98.30\pm0.20\%$ training and $75.97\pm0.75\%$ test accuracy, so the network does not quite fit its training set. A broad region of positive test gain appears [Fig.~\ref{fig:sm-cnn-grids}(a)]: $85$ of the $195$ DB configurations improve on the UB test accuracy, $42$ of them by more than one point. The best DB test accuracy is $77.96\pm0.60\%$ (at $A=75$, $T=1500$), a gain of $+2.00\pm0.96$ points, and several other cells come within $0.1$ points of it [$(A,T)=(50,1500)$, $(100,150)$ and $(100,400)$]. Training accuracy in the best configuration falls to $90.07\pm0.35\%$, and the highest DB training accuracy, $98.19\pm0.18\%$ (at $A=5$, $T=30$), is within noise of the baseline: as for the narrow ResNet-10, the protocol acts as a regularizer.
 
\subsection{Myrtle5}
 
A VGG-style Myrtle5 network~\citep{shankar2020neural}: four convolutional stages with one block each, channel widths doubling from $64$ up to $512$, $3\times3$ kernels, batch normalization, and pooling at the last stage.
 
\textbf{Results:} The UB baseline reaches $100.00\pm0.00\%$ training and $85.10\pm0.47\%$ test accuracy. Under DB, training accuracy stays at $100.00\%$ wherever training does not collapse, and the best test accuracy is $85.64\pm0.59\%$ (at $A=5$, $T=15$). The difference, $+0.53\pm0.75$ points, is smaller than its combined s.d., so we treat it as no effect. The distinction from the capacity-limited models rests on this ratio of gain to seed noise, not on how the configurations were selected, which is the same for all architectures: $+1.43$ points is about $2.3$ times its combined s.d.\ ($0.62$) and $+2.00$ about twice ($0.96$), whereas for the two wider networks the gain is below its combined s.d.
 
\subsection{ResNet-10 at half width}
 
The ResNet-10 of the main text, four stages of one basic block each, with width multiplier $0.5$ instead of $0.15$.
 
\textbf{Results:} The UB baseline reaches $100.00\pm0.00\%$ training and $85.98\pm0.48\%$ test accuracy. The best DB test accuracy is $86.20\pm0.07\%$ (at $A=10$, $T=50$), a difference of $+0.22\pm0.48$ points. No DB configuration improves test accuracy by more than half a point. Widening the same architecture from $0.15$ to $0.5$ thus raises the UB test accuracy by about six points and removes the benefit of the protocol.


\clearpage
\bibliography{references}

\end{document}